\pdfoutput=1

\documentclass[11pt]{article}

\usepackage{EMNLP}

\usepackage{times}
\usepackage{latexsym}
\usepackage{hyperref}
\usepackage{subfigure}
\usepackage{subcaption}
\usepackage{enumitem}
\usepackage{booktabs}
\usepackage{multicol}
\usepackage{multirow}
\usepackage{makecell}
\usepackage{amsmath,amssymb,amsthm}
\usepackage{mathtools,thmtools}
\usepackage{algorithm}
\usepackage{mdframed}
\usepackage{bm}
\usepackage{siunitx}
\usepackage{afterpage}
\usepackage{float}
\usepackage[T1]{fontenc}

\usepackage[utf8]{inputenc}

\usepackage{microtype}

\newcommand{\ie}{\textit{i}.\textit{e}.}
\usepackage{inconsolata}
\newcommand{\eg}{\textit{e}.\textit{g}.}
\renewcommand{\hat}[1]{\widehat{#1}}

\newmdtheoremenv{thm}{Theorem}
\newmdtheoremenv{rem}{Remark}
\newmdtheoremenv{appmode}{Use Case}

\title{Collaborative Performance Prediction for Large Language Models}

\author{
  Qiyuan Zhang \\
  City University of Hong Kong \\
  \texttt{qzhang732-c@my.cityu.edu.hk} \\\And
  Fuyuan Lyu\thanks{\; Co-corresponding Authors} \\
  McGill University \& MILA \\
  \texttt{fuyuan.lyu@mail.mcgill.ca} \\\AND
  Xue Liu \\
  McGill University \\
  \texttt{xueliu@cs.mcgill.ca} \\\And
  Chen Ma\footnotemark[1] \\
  City University of Hong Kong \\
  \texttt{chenma@cityu.edu.hk}
  }

\begin{document}
\maketitle
\begin{abstract}
Comprehensively understanding and accurately predicting the performance of large language models across diverse downstream tasks has emerged as a pivotal challenge in NLP research.
The pioneering scaling law on downstream works~\cite{hu2024predicting,isik2024scaling} demonstrated intrinsic similarities within model families and utilized such similarities for performance prediction.
However, they tend to overlook the similarities between model families and only consider design factors listed in the original scaling law.
To overcome these limitations, we introduce a novel framework, Collaborative Performance Prediction (CPP), which significantly enhances prediction accuracy by leveraging the historical performance of various models on downstream tasks and other design factors for both model and task. 
We also collect a collaborative data sourced from online platforms containing both historical performance and additional design factors.
With the support of the collaborative data, CPP not only surpasses traditional scaling laws in predicting the performance of scaled LLMs but also facilitates a detailed analysis of factor importance, an area previously overlooked.
Our code is available here\footnote{https://github.com/Don-Joey/CPP\_LLM}.
\end{abstract}

\section{Introduction}

\begin{figure*}[!htbp]
\centering
\includegraphics[width=0.98\textwidth]{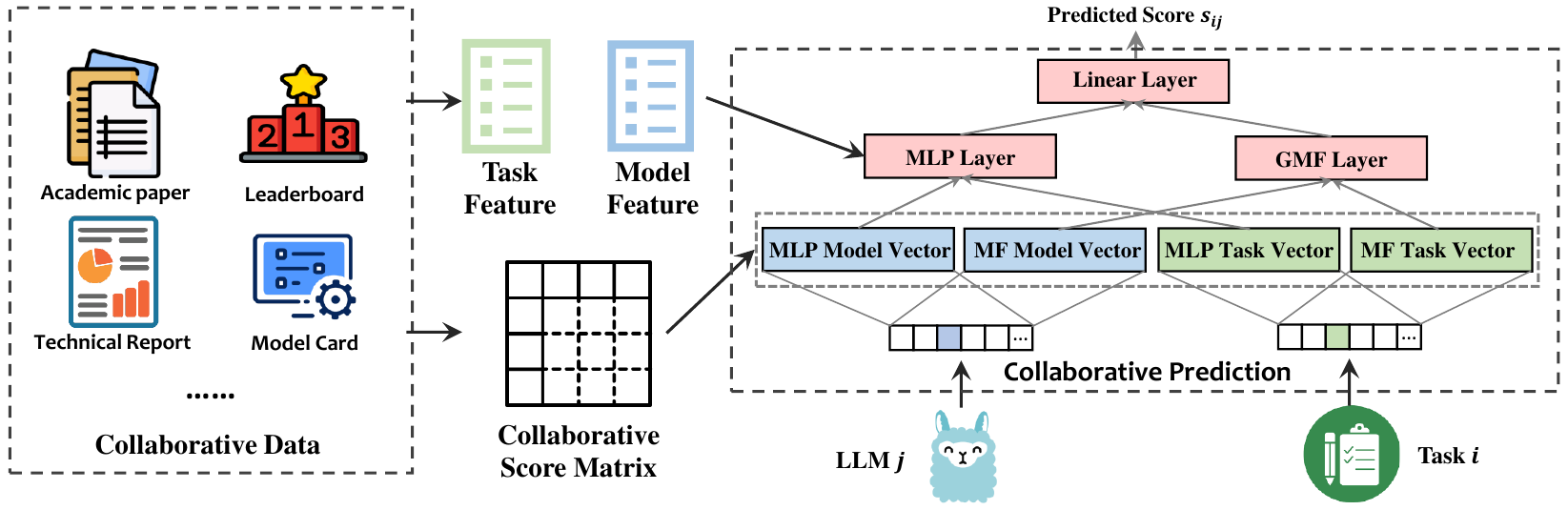}
\caption{Framework for Collaborative Performance Prediction of Large Language Models. This schematic delineates two principal components: (1) Collaborative Data, which encompasses a score matrix illustrating the performance of various LLMs across downstream tasks, along with external descriptive factors of both models and tasks; (2) Collaborative Prediction Method, given the model and task IDs to leverage this collaborative data, enabling accurate score prediction.}
\label{fig:framework}
\end{figure*}

Large Language Models (LLMs)~\cite{brown2020language,ouyang2022training} have emerged as one of the most important AI research powered by large-scale parameters, high computational resources, and massive training data. 
With the substantial increase in model sizes, the evaluation cost of LLMs' performance becomes even more significant.
For example, testing a single LLM on certain benchmarks often requires $\$10$K+ and $4$K+ GPU hours~\cite{liang2023holistic}.
Therefore, understanding the behaviors and predicting the capabilities of LLMs across scales under various tasks becomes a vital question~\cite{Ganguli2022,owen2024predictable,finnveden2020extra,hu2024predicting} for both researchers and engineers.

Scaling laws~\cite{kaplan2020scaling, hoffmann2022an, hernandez2022scaling, gordon2021data, bahri2024explaining, muennighoff2023scaling} have been powerful tools for predicting the capabilities of LLMs. 
It indicates a power-law correlation between the model performance and design factors such as computational measure (\textit{FLOPs}) utilized during training.
Although the scaling law was originally proposed as a strong intuitive guide for designing LLM, researchers~\cite{hu2024predicting, ruan2024observational, isik2024scaling} have extended its utility into predicting model performances on various metrics, such as BLEU in Machine Translation, and different tasks.
These works can accurately predict model performances by utilizing the similarity within each model family, \eg, models within each family are usually trained on the same dataset.
However, there are several issues rooted in their methods: the performance prediction 1) requires transparent design factors that consume substantial training resources to fit the curve, 2) is only tailored to a certain model family and a specific task metric, and 3) neglects the connections among different models and tasks.

The aforementioned limitations motivate us to design more effective methods for predicting the performance of LLMs on downstream tasks.
Two observations sparked our attention. 
Firstly, A strong similarity exists between model families, \eg LLama-family and GPT family. Models from different families behave similarly in prediction distribution~\cite{shrivastava2023llamas} and emergent phenomenon~\cite{wei2022emergent}. 
Secondly, with the emerging LLM models and the increasingly diverse tasks, the cost of enumerating and benchmarking models with tasks increases exponentially.
Therefore, we aim to utilize the similarities across model families in order to collaboratively predict the model performance over diverse tasks in an accurate yet efficient way.

To incorporate the aforementioned intuitions, we propose a new scheme, Collaborative Performance Prediction (\textsc{CPP}), to efficiently predict the performance of LLMs on evaluation tasks. This scheme learns the latent representations of LLMs and tasks, which captures the intrinsic similarity among different models and tasks. The interaction (\eg, inner product) between the latent representations of LLMs and tasks can be utilized to predict the performance of LLMs on certain tasks. To fulfil the proposed scheme, we collect the LLM performance data from academic papers, technical reports, and open leaderboards covering 72 models and 29 tasks. To summarize, our scheme has several advantages:
\begin{itemize}[topsep=0pt,noitemsep,nolistsep,leftmargin=*]
    \item \textbf{Low Training Cost}: Compared with methods~\cite{hu2024predicting} that extend scaling law to various downstream tasks, no pre-training or fine-tuning of LLM is required in our scheme. 
    \item \textbf{Prediction over proprietary model}: Unlike previous methods~\cite{ruan2024observational}, our scheme supports prediction over proprietary models without knowing key design factors, such as computational measures.
    \item \textbf{Prediction from small to large:}
    By utilizing cross-family information, our scheme can accurately estimate model performance, \eg, emergent ability, of large models on downstream tasks given the information from small models.
    \item \textbf{Beyond Scaling Laws}: Our scheme is more general and can incorporate diverse factors, such as task description factors.
    \item \textbf{Factor-level Interpretability}: Our scheme can provide interpretability by analyzing the factors importance of LLMs.
\end{itemize}

Under our scheme, multiple customized prediction methods (\eg, \textsc{Collaborative Fitering}~\cite{Koren2022}) can be incorporated to predict the performance of LLMs, further validating the feasibility and generality. 
Our method enables more diverse factors as input, ranging from traditional LLM design factors to task design factors, \eg, targeted ability and few-shot setting.

Upon extensive experimentation within the open-released core leaderboard of HELM~\cite{liang2023holistic} and our collected historical matrix, our predictive performance demonstrated exceptionally well. Specifically, even without any input of model factors or task factors: in HELM, we use 50\% of the scores to predict the other 50\%, the predictive ranking (derived from predicted scores) achieves $Accuracy=$10\%, and $MAE@2=$39\%; in our collected matrix (characterized by a 44\% sparsity level) achieves an $Accuracy=$45\%, and the $MAE@2=$84\%. Notably, the accuracy of our prediction from small to large LMs significantly exceeded that predicted by scaling laws.
Using an analysis method similar to \textsc{Shapley-Values}~\cite{lundberg2017unified,shapley}, we elucidate the importance of different factors, which surprisingly does not fully align with scaling law~\cite{kaplan2020scaling}.
Therefore, our method is undoubtedly more versatile.

\section{Related Work}

\subsection{Downstream Scaling Law and Performance Predictability of LLM}

Scaling laws~\cite{kaplan2020scaling,hoffmann2022an,hernandez2022scaling,bahri2024explaining,muennighoff2023scaling} for LLMs have increasingly become a focal point in understanding and guiding critical design decisions, such as model size and the characteristics and volume of pretraining data. Traditionally, most research in this area has concentrated on how measures like cross-entropy loss or perplexity scale. Subsequent studies have extended these efforts to the scaling behavior on translation~\cite{isik2024scaling,ghorbani2021scaling,zhuocheng2023scaling} and other downstream tasks modeling~\cite{caballero2023broken,henighan2020scaling}. 
The high predictability in LLMs capability has directly spurred extensive research work (see Survey~\citet{anwar2024foundational}) exploring whether LLMs can demonstrate predictability on downstream tasks, which are considered highly unpredictable in traditional ML knowledge~\cite{Ganguli2022}. Particularly, the ``emergence'' phenomenon~\cite{suzgun2022challenging,wei2022emergent} has challenged predictability, where models suddenly exhibit striking capabilities at specific training reources. Recent studies~\cite{schaeffer2023are} have made remarkable achievements in breaking the discontinuities in performance brought about by emergence, and \citet{Ganguli2022,owen2024predictable,finnveden2020extra} demonstrated the predictability on downstream tasks, for instance, \citet{hu2024predicting} directly fits a curve of training resources and downstream task performance by repeatedly pretraining a specific model. Furthermore, \citet{arora2023theory} predicted the performance through decomposing the complex capabilities of LMs to some base skills.

Given that predictability has now been established, we reassess the underlying premises that enable this predictability: the  prevailing similarities across multiple models and various downstream tasks~\cite{liu2023question,perlitz2024efficient,polo2024tinybenchmarks,torregrossa2020correlation,ilić2023unveiling}. Based on this, we step beyond the limitations defined by scaling laws and propose a new methodology to predict the performance of LLMs on various downstream tasks.

\subsection{Collaborative Filtering} \label{sec:cf}

Collaborative filtering (\textsc{CF})~\cite{Koren2022} is a widely used technique in recommendation systems that predicts users' preferences by collecting the historical preferences of many other users. The underlying assumption of \textsc{CF} is that similar users will share similar preferences on similar items. A seminal method in \textsc{CF} is matrix factorization~\cite{koren2009matrix} (\textsc{MF}). It reduces the dimensionality of the user-item matrix by learning the latent factors associated with users and items, respectively. This approach helps handle sparse data and improves scalability. The factorization of the user-item matrix $\mathbf{R}$ can be represented as
\begin{equation}
    \mathbf{R} \approx \mathbf{P}^\top \cdot \mathbf{Q} \,,
\end{equation}
where each column vector in $\mathbf{P}$ and $\mathbf{Q}$ represents a specific user or item, respectively, with hidden dimension $ d $. The latent representations of users and items capture the user preferences and item properties in the latent space, and the inner product $ \cdot $ can be utilized to predict the interaction between users and items. To optimize the latent feature vectors, the following loss function is employed:
\begin{equation}
    \min_{\mathbf{P}, \mathbf{Q}} \quad \sum_{(u, i) \in \Omega} (r_{ui} - \mathbf{p}_u^\top \cdot \mathbf{q}_i)^2\,,
\end{equation}
which measures the squared differences between the observed ratings $r_{ui}$ and the ratings predicted by the model $\mathbf{p}_u^\top \cdot \mathbf{q}_i$ for each user-item pair $(u, i)$ in the set $\Omega$ of observed interactions. 

Here, \citet{yang2019oboe} transferred the collaborative filtering for ML model selection by predicting the cross-valided errors,  which demonstrates \textsc{CF}'s adaptability and efficiency in diverse application areas.

\section{Background and Pilot Demonstration}
\begin{figure*}[ht]
\begin{center}
\centerline{\includegraphics[width=0.98\textwidth]{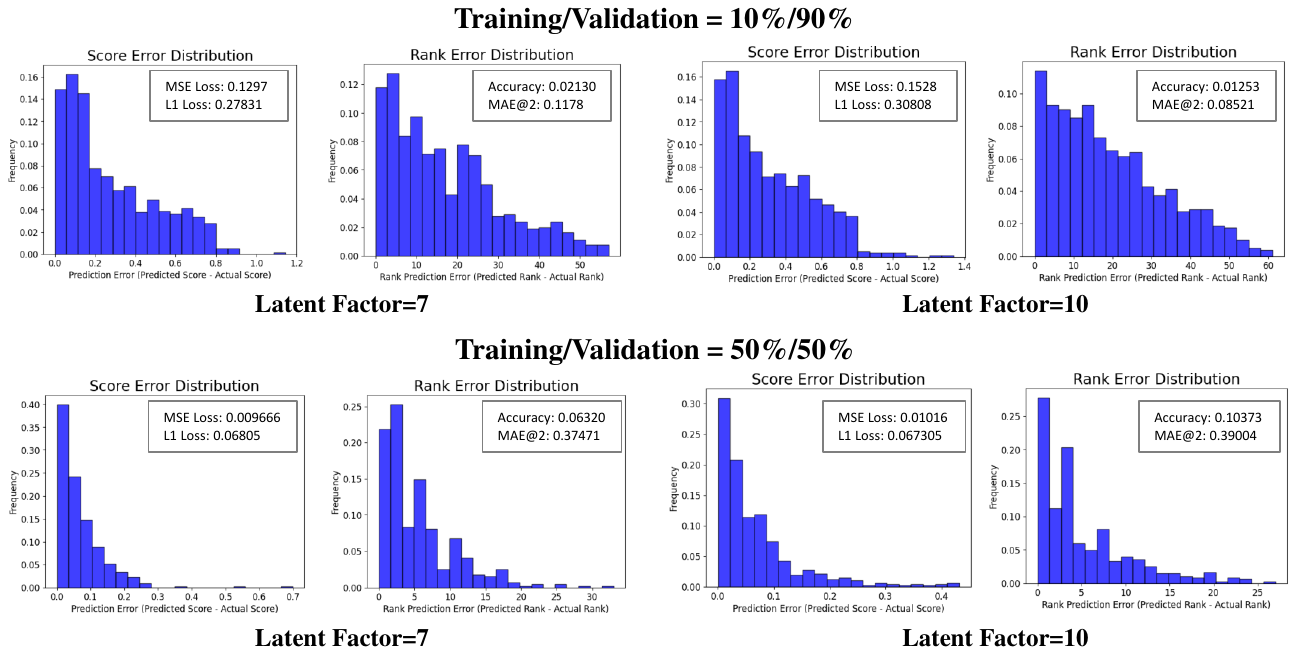}}
\caption{\textbf{Error Distribution of Predictions (Normalized Score and Rank Derived by Score) Based on the HELM Lite Leaderboard Using Matrix Factorization:} We evaluate the effectiveness of Matrix Factorization (MF) using two latent factors, 7 and 10, across 2 training/validation split percentages. \textbf{Accuracy} is the percentage of instances where the predicted rank equals the actual rank. \textbf{MAE@2} is defined as the percentage of instances where the absolute difference between the predicted and actual ranks is 2.
}
\label{fig:pilot}
\end{center}
\end{figure*} 
\subsection{Scaling Law on Downstream Tasks}

For classic scaling laws, researchers propose a hypothesized power-law relationship between a model's computational measures $C_m$ (\eg, training FLOPs) and their performance loss $L_m$ (\eg, perplexity). Specifically, for a model $m$ within a family $f$ (\eg, Llama-2 7B, 13B, and 70B), the relationship is hypothesized as
\begin{equation}
    \log(L_m) \approx \omega_f \log(C_m) + \ b_f \,,
\end{equation}
where $\omega_f$ and $\ b_f$ are scaling coefficients customized for each model family. Researchers fit this formula through repeated scaling experiments, then use it to accurately predict performance when larger-scale ($C' > C$). Some studies~\cite{finnveden2020extra,owen2024predictable} have adapted scaling laws to specific downstream task metrics, proposing that sigmoidal functions are more suitable for predictions, as follows:
\begin{equation}
    \sigma^{-1}(S_m) \approx \omega_f\log(C_m) + \ b_f,
\end{equation}
where $S_m$ refers to the normalized downstream scores of models within the range $[0, 1]$. However, applying scaling laws across different model families on various specific tasks presents a trade-off: fitting unique coefficients for each evaluation scenario (\eg, Llama 2 on \textsc{MMLU}) is a resource-intensive endeavor~\cite{hu2024predicting}; alternatively, estimating these coefficients using a limited number (3-5) of models within the same family may compromise the accuracy of the predictions. Moreover, the recent work~\cite{ruan2024observational} extends scaling law by incorporating latent variables to capture the patterns across model families and tasks.

\subsection{Pilot Demonstration on HELM}
Scaling laws reveal that models from any family exhibit a similar performance trend as computational measures increase. This insight suggests there are commonalities and connections between different models. These motivate us to employ the \textsc{MF} method to explore more similarities beyond computational measures, \eg, the relationship among the different model families and tasks.

We perform the aforementioned \textsc{MF} on the benchmark matrix to observe the error gap between predicted and truth (normalized) scores. Specifically, we select the core leaderboard provided by \textsc{HELM} for our exploratory experiments with only model name, task name, and performance scores. This leaderboard, 68 models and 16 tasks, presented in a score matrix with a density of $82.5\%$, which includes both open-source and proprietary models, \eg, GPT-4 and Jurassic-2. Our method treats all models and tasks as independent entities without introducing any prior similarity factors. We hope to observe whether \textsc{MF} can predict the remaining scores, giving a small part of the matrix, where we evaluate two training/validation sets split strategies: 10\%/90\%, 50\%/50\%. As illustrated in Figure~\ref{fig:pilot}, \textsc{MF} can accurately predict most of the missing scores within a low error range, which proves that it can encode the similarity across the model and the task without regression depending on explicit computational measures variable.

\section{Collaborative Performance Prediction}

\subsection{Definition}

Motivated by pilot experiments, we introduce the concept of ``Collaborative Performance Prediction'' (CPP) to facilitate the performance prediction of LLMs.

\newtheorem{definition}{Definition}
\begin{definition}
Let $\mathcal{M} = \{M_1, M_2, \ldots, M_n\}$ be a set of $n$ LLMs, and $\mathcal{T} = \{T_1, T_2, \ldots, T_m\}$ be a suite of $m$ tasks. Define the Score Matrix $\mathbf{S}$, which is an $n \times m$ matrix where each element $s_{ij}$ represents the performance score of model $M_i$ on task $T_j$. $s_{ij}$ is defined as
        \[
        s_{ij} = \begin{cases} 
        \text{score}  & \text{if tested,} \\
        \text{unknown} & \text{otherwise}.
        \end{cases}
        \]
    \textbf{Function:} Employ an prediction method $\mathcal{F}$ to estimate the unknown elements of $\mathbf{S}$, denoted by $\hat{s}_{ij}$, based on the known values.\\
    \textbf{Extention:} Accommodate model design factors $\mathcal{V}_m = \{V_m^1, V_m^2, \ldots, {V_m^M}\}$, such as common computational meatures, and task design factors $\mathcal{V}_t = \{V_t^1, V_t^2, \ldots, {V_t^T}\}$, such as targeted capabilities and few-shot settings.
\end{definition}

Based on this definition, our framework consists of two components: 1) \textbf{collaborative performance data}, 2) \textbf{collaborative prediction methods}. We anticipate that an accurate score can be predicted based on the historical performance of various models on downstream tasks and other design factors for both model and task. Moreover, we can incorporate or solely rely on the factors describing the LLM and the associated downstream tasks.

\subsection{Collaborative Data}
\begin{figure}[htbp]
\begin{center}
\centerline{\includegraphics[width=0.48\textwidth]{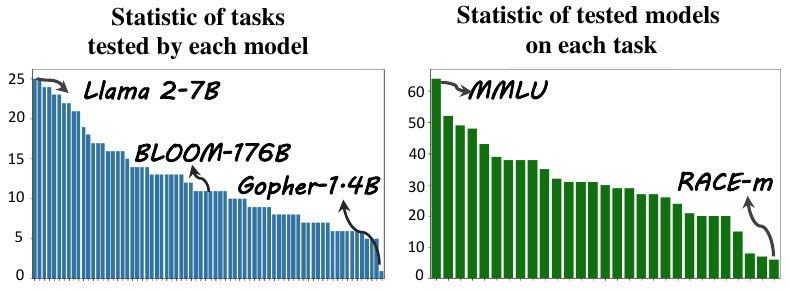}}
\caption{Distribution of Testing Coverage Across Models and Tasks. The left bar shows the number of tasks each model has been tested on; The right bar illustrates the number of models tested in each specific task.}
\label{fig:data_analysis}
\end{center}
\end{figure}
Unlike the scaling law approach, which requires training resource factors to obtain the correlation between metric scores and factors at a high training cost, our proposed method makes use of evaluation results and other design factors reported from existing studies, referred to as \textit{collaborative data}. 
Open-source leaderboards, such as OpenLLM\footnote{\url{https://github.com/bentoml/OpenLLM}}, HELM, and OpenCompass\footnote{\url{https://opencompass.org.cn/}}, have made tremendous efforts on this issue in fairly evaluating different LLMs.
Our efforts extend beyond merely~\cite{ruan2024observational} utilizing data from open-source leaderboards with matrix sparsity of 0\%. 
We also extract test results from different models' papers, technical reports, and model cards. 
Ultimately, we have collected a score matrix of $n=72$, $m=29$ with a density of only $56\%$. 
Furthermore, we collected 12 and 4 detailed design factors for models and tasks. 
These details are listed in Appendix~\ref{sec:description_features}. 
Our data analysis is shown in Figure~\ref{fig:data_analysis} and Figure~\ref{fig:detailed_analysis}.

\paragraph{Data Analysis.} 
Based on the \textit{collective data}, we can make the following observations: 
a) \textbf{Uneven distribution of testing resources}. 
We observe significant variability in the deployment of testing efforts, as shown in Figure~\ref{fig:data_analysis}. 
For instance, models from the \textsc{Llama} series have undergone extensive testing across various tasks, in contrast to models like \textsc{Gopher}, where testing has largely stagnated. 
A similar disparity is also evident among tasks, with \textsc{MMLU} and \textsc{Hellaswag} receiving considerable evaluation, whereas \textsc{RACE} has been relatively underexplored. 
This trend suggests that as LLMs proliferate and tasks evolve, scores across the matrix will increasingly skew. This leads to a pronounced long-tail effect in testing coverage for many tasks, barring a few that consistently receive comprehensive evaluations.
b) \textbf{Widespread variations in the scores}. 
It is noteworthy that identical models yield varying scores on the same tasks across different studies~\cite{shrivastava2023llamas,llama3modelcard}, a variation often attributed to differences in prompt settings, model versions, and the volume of test samples employed. 
Typically, these score variations range within 0.1, with scores normalized between $[0,1]$. This phenomenon underscores the importance of public leaderboards and highlights researchers' need to articulate their testing frameworks when performing customized evaluations. 
When conflicted, we prefer the results from the Open-LLM leaderboard in the collective data.
c) \textbf{Missing description/model card}. 
We advocate for consistently providing complete model cards for open-source and proprietary models.
Such a phenomenon is shown in Figure~\ref{fig:detailed_analysis} and, unsurprisingly, a long-tail distribution is witnessed.
While it is understandable that proprietary models might withhold specific details about parameters, they can still divulge information about parameter scale and the extent of pre-training. 
Furthermore, we recommend a more thorough description of testing tasks, including suggested few-shot settings and detailed descriptions of targeted capabilities.

\subsection{Prediction Methods}

In Section~\ref{sec:cf}, classical collaborative filtering methods are inspired to conduct the performance prediction. In principle, most collaborative filtering methods can be applied. Here, in addition to the abovementioned \textsc{MF}, we also leverage neural collaborative filtering~\cite{he2017neural} (\textsc{NCF}) methods, which uses a multi-layer perceptron to learn the model-task interaction function to predict the score $\hat{s}_{ij}$ for a model $i$ on a task $j$, providing a way to learn non-linearities in the data:
\begin{equation}
\begin{aligned}
    \hat{s}_{ij} &= f(i, j | \mathcal{M}, \mathcal{T}, [\mathcal{V}_i, \mathcal{V}_j], \theta)\\
    &= \text{MLP}(p_i, q_j, [e_{vi}, e_{vj}]),
\end{aligned}
\end{equation}
where $\mathcal{M}$ and $\mathcal{T}$ denote the sets of collaborative models and tasks, and their descriptive factors $\mathcal{V}_i$, $\mathcal{V}_j$ optionally enrich the input data. Here,
\( p_j \) and \( q_j \) are the latent vectors for model \( i \) and task \( j \) that capture the intrinsic properties of models and tasks, as well as embeddings \( [e_{vi}, e_{vj}] \) derived from their descriptive factors, and \( \theta \) represents the parameters of \textsc{NCF}.

Moreover, we further simplify the model to verify whether it is feasible to predict a score when only inputting the descriptive factors $\mathcal{V}_i$, $\mathcal{V}_j$ into the prediction model:

\begin{equation}
\begin{aligned}
    \hat{s}_{ij} &= f(i, j | \mathcal{V}_i, \mathcal{V}_j, \theta)\\
    &= \text{MLP}(e_{vi}, e_{vj}),
\end{aligned}
\end{equation}

For both settings, where the goal is to predict the scores accurately, the loss function can be defined as follows:
\begin{equation}
L(\theta) = \frac{1}{N} \sum_{(i, j) \in \mathcal{D}} (\hat{s}_{ij} - s_{ij})^2,
\end{equation}
where $N$ is the total number of scores set $\mathcal{D}$ for training, and $s_{ij}$ is the true score for model \( i \) and task \( j \).
\section{Experiments}
\label{sec:validation}

In this section, we evaluate the feasibility of \textsc{CPP} from an overall benchmark perspective and a model perspective in Section~\ref{sec:exp_overall} and ~\ref{sec:exp_model}, respectively;  we then analyze the importance of factors for both models and tasks in Section~\ref{sec:feature_importance}. Additionally, a substantial amount of ablation and analysis is placed in the appendix~\ref{sec:ablation}, such as sparsity, the correlations in tasks and models, and which models and tasks are more critical for prediction.

\paragraph{Experimental Setting.} 
Our validation framework utilizes the aforementioned collaborative dataset as the score matrix $\mathcal{S}$. 
We partition scores $\{s_{ij}\}$ into train and validation set, detailed in Appendix~\ref{sec:validation_setting}.

\paragraph{Evaluation Metric.}
To accurately evaluate \textsc{CPP}, we adopt two types of metrics: 1) \textsc{Score-Loss} metrics including \textsc{MSE Loss} and \textsc{L1 Loss} between predicted scores and true scores (normalized) on downstream tasks and 2) \textsc{Rank-Accuracy} metrics including \textsc{Accuracy} and \textsc{MAE@2} between the rank of predicted scores and true scores. We elaborate on these metrics in Appendix~\ref{sec:appendix_evaluation}.

\subsection{Evaluation from Benchmark Perspective} 
\label{sec:exp_overall}
\begin{figure*}[!t]
\centering
\includegraphics[width=0.95\textwidth]{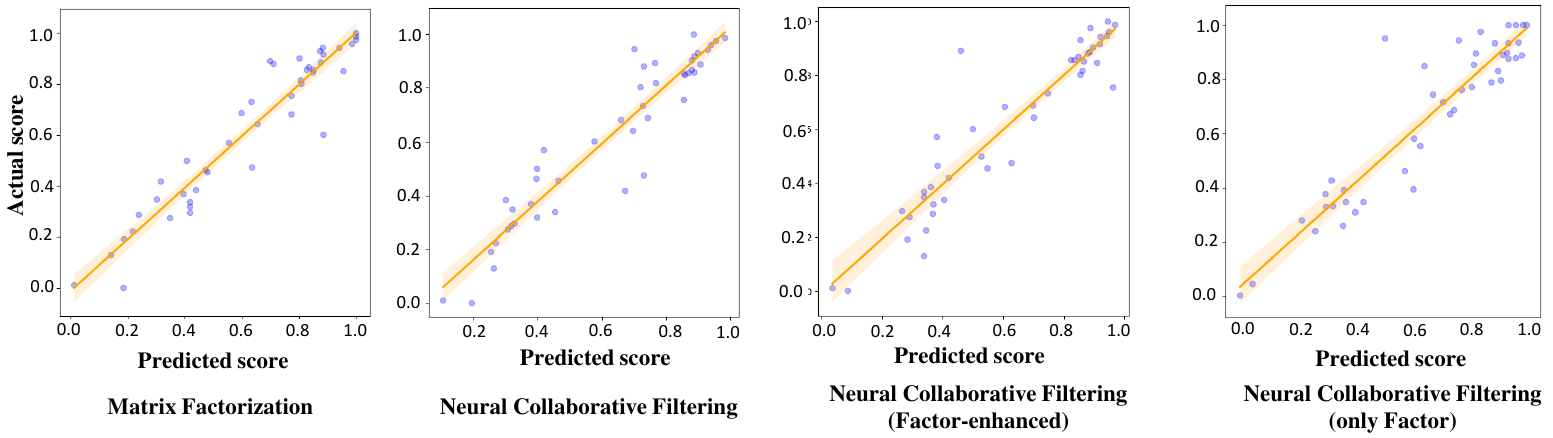}
\caption{Comparative visualization of predictive accuracy across various scoring methods. From left to right: MF, NCF, NCF with Factor Enhancement, and NCF based solely on Factors. Each plot displays the regression between predicted and actual scores, where the solid line represents the regression fit and the shaded area denotes the confidence interval (CI). A line closer to the diagonal indicates perfect prediction and higher prediction accuracy. These plots demonstrate the enhanced performance in score prediction achieved by integrating factors into the NCF method.}
\label{fig:different_models}
\end{figure*}

\begin{table*}[!t]
\resizebox{0.98\textwidth}{!}{
\centering
\begin{tabular}{lcccc} 
\hline
\multirow{2}{*}{\textbf{Prediction Method} }& \multicolumn{2}{c}{\textbf{Score-Loss}} & \multicolumn{2}{c}{\textbf{Rank-Acc}}\\
\cline{2-3} \cline{4-5}
& MSE Loss $\downarrow$ & Mean L1 Loss $\downarrow$ & Mean Prec.(\%) $\uparrow$ & MAE@2(\%) $\uparrow$ \\
\hline 
Matrix Factorization            & $2.16e^{-2}$\text{\small($1.19e^{-4}$)} & $9.47e^{-2}$\text{\small($2.89e^{-4}$)} & $44.33$\text{\small($0.69$)} & $83.16$\text{\small($0.73$)} \\
Neural Collaborative Filtering  & $1.58e^{-2}$\text{\small($4.22e^{-5}$)} & $8.94e^{-2}$\text{\small($3.10e^{-4}$)} & $41.76$\text{\small($1.22$)} & $\mathbf{84.98}$\text{\small($\mathbf{0.42}$)} \\
\qquad\qquad + Factor Enhanced & $\mathbf{1.25e^{-2}}$\text{\small($\mathbf{3.35e^{-6}}$)} & $\mathbf{7.88e^{-2}}$\small($\mathbf{6.31e^{-5}}$) & $\mathbf{45.45}$\small($\mathbf{0.33}$) & $84.54$\text{\small($0.27$)} \\
\hline
Only Factor                    & $1.75e^{-2}$\text{\small($2.07e^{-5}$)} & $8.57e^{-2}$\text{\small($1.48e^{-4}$)} & $33.47$\text{\small($0.12$)} & $84.08$\text{\small($0.37$)} \\
\hline
\end{tabular}
}
\caption{Comparison of prediction methods for LLM performance. \textbf{Bold} indicates the best-performed.}
\label{tab:differ_models}
\end{table*}

In this study, we select the abovementioned methods, MF and NCF, to verify whether $s_{ij}$ can be accurately predicted based on the input of model \( i \) and task \( j \). 
To examine whether enhancements are helpful, we modify NCF to support the input of design factors, detailed in Appendix~\ref{sec:validation_setting}.
Based on Figure~\ref{fig:different_models} and Table~\ref{tab:differ_models}, we can make the following observations:

First, all methods accurately predicted model performance, demonstrating that collaborative filtering mechanisms can predict model outcomes based on collaborative data across different models and tasks. 
This prediction is achieved without explicit scaling factors or fitting a log-power curve. 
Second, from \textsc{MF} to \textsc{NCF}, the transformation in interaction mechanisms further enhances accuracy, suggesting that model improvements can further augment the efficacy of our methodology. 
Additionally, we further increased accuracy by incorporating factors, such as model scaling variables and task descriptions, into the NCF framework alongside ID information. 
This confirms that incorporating explicit factors can enhance model and task similarities. 
Finally, among all metrics, we particularly noted that the accuracy of the predictive ranking was acceptable. 
In other words, researchers can use our method to accurately predict the ranking range of their developed models on test tasks, thereby enhancing model performance on specific tasks.

\paragraph{Predictability with Only Description Factors.} 
We validate whether high predictive accuracy can still be achieved by only inputting the models' and tasks' design factors. 
As demonstrated in Table~\ref{tab:differ_models}, the accuracy of predicted rankings (derived from predicted scores) remains high, affirming that our method supports predictions based solely on factors. 
However, the accuracy is lower than other models, suggesting that finer-grained latent similarities remain encoded as potential factors within the identity information across different models and tasks.

\subsection{Evaluation from Model Perspective}
\label{sec:exp_model}
\begin{figure}[!t]
\centering
\includegraphics[width=0.43\textwidth]{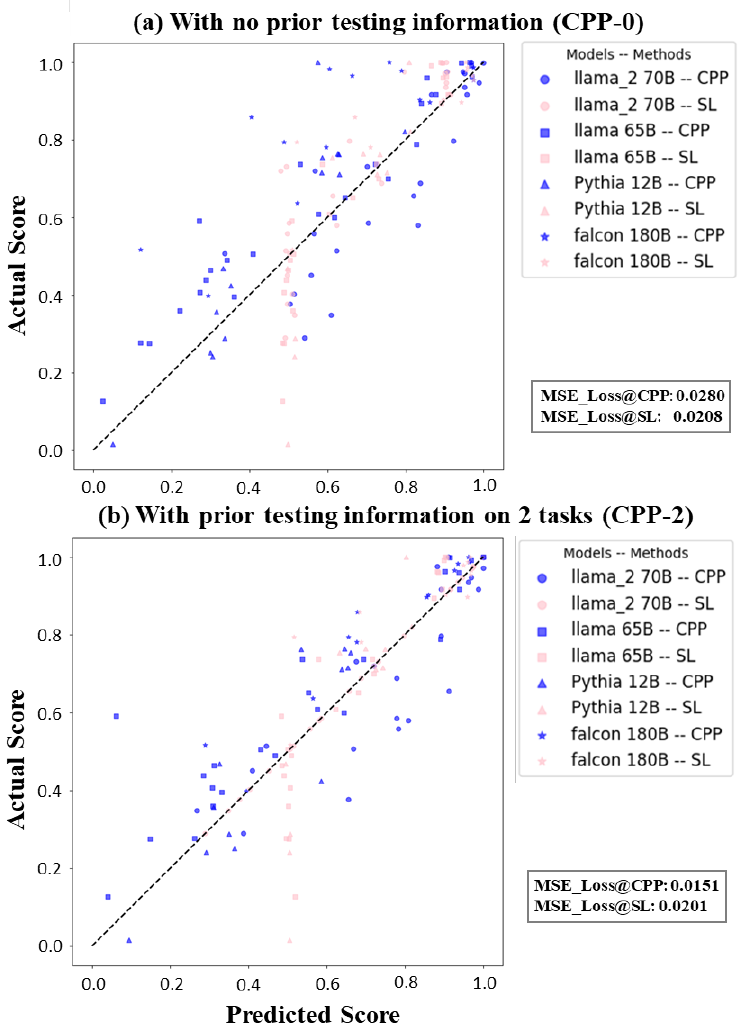}
\caption{Comparison of the predictive performance of collaborative performance prediction (CPP) versus traditional scaling laws (SL) for LLMs: (a) CPP-0, with no prior testing information, and (b) CPP-2, with prior testing on two tasks.}
\label{fig:predicting_llm}
\end{figure}

To mimic the utilization of \textsc{CPP} in the real world, this section takes a model perspective to investigate the predictive accuracy of \textsc{CPP} upon each model. 
Specifically, we propose two scenarios: (i) prediction with no prior testing information and (ii) prediction with prior testing information on 2 tasks. These two scenarios correspond to real-world cases when the model has not been developed or is tested on a few tasks and expects an accurate prediction of its ability on other tasks. In both scenarios, we focus on larger LLMs, \eg, LLama2-70b, as they are more computationally expensive to develop and test, requiring an accurate LLM prediction.

We report the results of \textsc{CPP} and \textsc{SL} on both scenarios in Figure~\ref{fig:predicting_llm} and can draw the following conclusions.
Under the \textit{CPP-0} scenario, \textsc{CPP} demonstrated greater adaptability across different tasks compared to \textsc{SL}, with points closely aligned along the $y=x$ line (``perfect prediction'') in Figure~\ref{fig:predicting_llm} (a). This suggests that \textsc{CPP} has effectively captured task-specific characteristics, such as value ranges, whereas \textsc{SL}, despite achieving a lower \textsc{MSE-loss}, tends to concentrate its predictions around 0.5.
Under the \textsc{CPP-2} scenario, the distribution of points of \textsc{CPP} is noticeably closer to $y=x$, as shown in Figure~\ref{fig:predicting_llm} (b), and its \textsc{MSE-loss} is also lower than that of \textsc{SL}. This indicates that leveraging performance data from other tasks considerably enhances the model’s cross-task prediction capabilities, underscoring a degree of consistency across tasks for the same model. This approach demonstrates that predictions for scaling LLMs on downstream tasks can be dynamically improved by evaluating performance on less computationally intensive tasks and using those outcomes to predict scores on subsequent tasks more accurately.

\subsection{Factor Importance Analysis via \textsc{Shapley-value}}
\label{sec:feature_importance}

\begin{figure}[!t]
\begin{center}
\centerline{\includegraphics[width=0.43\textwidth]{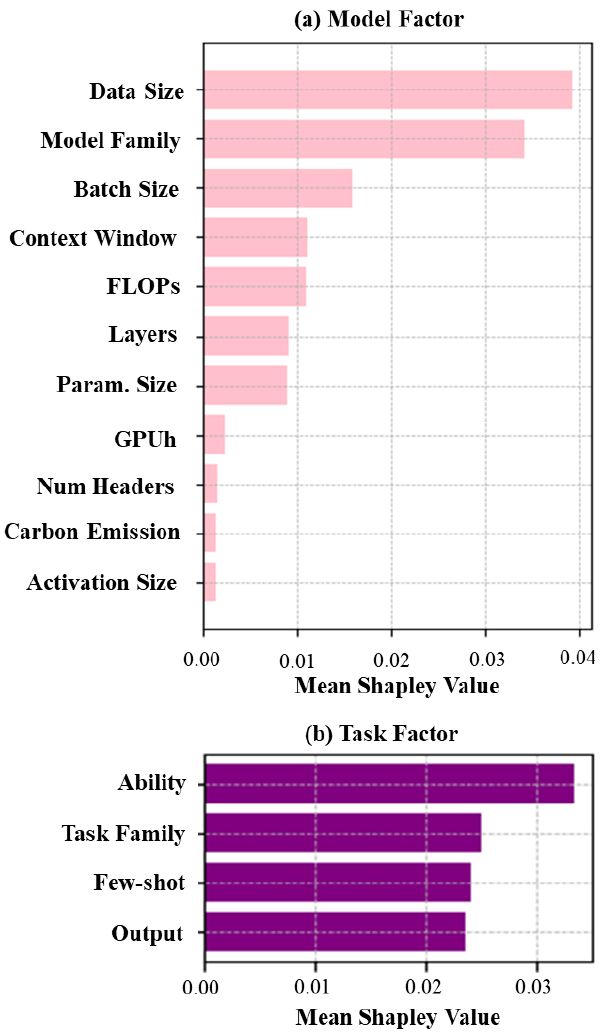}}
\caption{Mean Shapley Value on Each Factor.}
\label{fig:shapley}
\end{center}
\end{figure}

In this section, we aim to analyze each design factor's importance over \textsc{CPP}.
The Shapley value, a concept derived from cooperative game theory~\cite{shapley}, offers a systematic approach to measuring individual factors' contribution in predictive models~\cite{lundberg2017unified,covert2021explaining}. 
Appendix ~\ref{sec:validation_setting_shapley} shows a detailed formulation of the Shapley value.
Visualization for Shapley values of each design factor is shown in Figure~\ref{fig:shapley}.

Based on Figure~\ref{fig:shapley} (a), we can make the following observations regarding model factors.
First, we have discovered that in addition to traditionally important factors such as training data size and parameter size mentioned in scaling law~\cite{kaplan2020scaling}, other design factors significantly influence predictive outcomes. These include the model family, context window size, and batch size.
Second, the importance of the model family cannot be overlooked, as it may relate to differences in data quality across models, including proprietary data or specific architectural details. For instance, using a particular model family might mean adopting architectures or optimization techniques better suited to specific tasks. 
Moreover, the size of the context window also significantly affects model performance. A larger context window allows the model to better understand the context in long texts, which is particularly crucial for long-context LLMs~\cite{xiong2023effective}. Experience~\cite{gemini15} has shown that such models perform better across various tasks. Batch size, as another crucial factor, affects the stability and speed of model training. An appropriate batch size ensures a balance between the accuracy of gradient estimation and computational efficiency during training.

As for the importance of task factors, results in Figure~\ref{fig:shapley} (b) show that the \textit{target ability} among all factors is more important. This also implies that similarities between the domains of different tasks can help predict outcomes. This conclusion is consistent with previous observations ~\cite{ruan2024observational,perlitz2024efficient,polo2024tinybenchmarks}

In summary, these findings indicate that LLMs performance prediction should not rely solely on traditional design factors limited by scaling law but also on other key factors that might impact overall model performance.

\section{Conclusion and Discussion}
Advancing beyond traditional scaling laws on downstream tasks, we propose a collaborative performance prediction framework for large language models. It offers significant advantages, including easy deployment, low training costs, and superior predictive accuracy. Uniquely, it enables incorporating additional design factors and supports an in-depth analysis of their impact, including factor importance and correlations in models and tasks. For prediction, we collect collaborative data containing many historical performances and factors.

Our method's predictive accuracy is expected to improve as it benefits from an expanding pool of collaborative data. Moreover, this approach highlights the potential to identify neglected but vital factors beyond traditional scaling laws, such as task design factors, thereby enriching our comprehension of LLM performance predictability on downstream tasks.
\section*{Limitations}
\paragraph{``Single-source-of-truth''.} When collecting the \textit{collaborative data}, we hypothesize that each model's performance on each task is identical. However, in the real world, the detailed testing setting, for instance, the testing prompt writing, can influence LLM's performance variance. Although we observed this, we only saved one score from different sources. How to incorporate the setting of testing as an additional dimension remains to be solved in future works.
\paragraph{Susceptibility to data quality.} The prediction accuracy of \textsc{CPP} highly depends on the quality of collaborative data. The current version passively collects \textit{collaborative data} from online resources. Should information from either of these data sources be incorrect, the prediction capability of \textsc{CPP} would decrease correspondingly. To overcome such a limitation, jointly considering passive information collected from data sources and active information, such as performances of models tested on some tasks by the user, might be a solution. Utilizing techniques such as efficient benchmarking~\cite{perlitz2024efficient,polo2024tinybenchmarks} could alleviate the cost of obtaining active information.

\section*{Ethics Statement}
The data we use are collected from public papers, technical reports, open leaderboards, and model cards on GitHub.

\section*{Acknowledgements}
This work was supported by the Start-up Grant (No. 9610564), the Donations for Research Projects (No. 9229129) of the City University of Hong Kong, and the Early Career Scheme (No. CityU 21219323) of the University Grants Committee (UGC).

\bibliography{emnlp}

\begin{thebibliography}{49}
\expandafter\ifx\csname natexlab\endcsname\relax\def\natexlab#1{#1}\fi

\bibitem[{AI@Meta(2024)}]{llama3modelcard}
AI@Meta. 2024.
\newblock \href {https://github.com/meta-llama/llama3/blob/main/MODEL_CARD.md}
  {Llama 3 model card}.

\bibitem[{Anwar et~al.(2024)Anwar, Saparov, Rando, Paleka, Turpin, Hase,
  Lubana, Jenner, Casper, Sourbut, Edelman, Zhang, Günther, Korinek,
  Hernandez-Orallo, Hammond, Bigelow, Pan, Langosco, Korbak, Zhang, Zhong,
  hÉigeartaigh, Recchia, Corsi, Chan, Anderljung, Edwards, Bengio, Chen,
  Albanie, Maharaj, Foerster, Tramer, He, Kasirzadeh, Choi, and
  Krueger}]{anwar2024foundational}
Usman Anwar, Abulhair Saparov, Javier Rando, Daniel Paleka, Miles Turpin, Peter
  Hase, Ekdeep~Singh Lubana, Erik Jenner, Stephen Casper, Oliver Sourbut,
  Benjamin~L. Edelman, Zhaowei Zhang, Mario Günther, Anton Korinek, Jose
  Hernandez-Orallo, Lewis Hammond, Eric Bigelow, Alexander Pan, Lauro Langosco,
  Tomasz Korbak, Heidi Zhang, Ruiqi Zhong, Seán~Ó hÉigeartaigh, Gabriel
  Recchia, Giulio Corsi, Alan Chan, Markus Anderljung, Lilian Edwards, Yoshua
  Bengio, Danqi Chen, Samuel Albanie, Tegan Maharaj, Jakob Foerster, Florian
  Tramer, He~He, Atoosa Kasirzadeh, Yejin Choi, and David Krueger. 2024.
\newblock \href {http://arxiv.org/abs/2404.09932} {Foundational challenges in
  assuring alignment and safety of large language models}.
\newblock In \emph{arXiv}.

\bibitem[{Arora and Goyal(2023)}]{arora2023theory}
Sanjeev Arora and Anirudh Goyal. 2023.
\newblock \href {http://arxiv.org/abs/2307.15936} {A theory for emergence of
  complex skills in language models}.
\newblock In \emph{arXiv}.

\bibitem[{Austin et~al.(2021)Austin, Odena, Nye, Bosma, Michalewski, Dohan,
  Jiang, Cai, Terry, Le, and Sutton}]{austin2021program}
Jacob Austin, Augustus Odena, Maxwell Nye, Maarten Bosma, Henryk Michalewski,
  David Dohan, Ellen Jiang, Carrie Cai, Michael Terry, Quoc Le, and Charles
  Sutton. 2021.
\newblock \href {http://arxiv.org/abs/2108.07732} {Program synthesis with large
  language models}.
\newblock In \emph{arXiv}.

\bibitem[{Bahri et~al.(2024)Bahri, Dyer, Kaplan, Lee, and
  Sharma}]{bahri2024explaining}
Yasaman Bahri, Ethan Dyer, Jared Kaplan, Jaehoon Lee, and Utkarsh Sharma. 2024.
\newblock \href {http://arxiv.org/abs/2102.06701} {Explaining neural scaling
  laws}.
\newblock In \emph{arXiv}.

\bibitem[{Brown et~al.(2020)Brown, Mann, Ryder, Subbiah, Kaplan, Dhariwal,
  Neelakantan, Shyam, Sastry, Askell, Agarwal, Herbert-Voss, Krueger, Henighan,
  Child, Ramesh, Ziegler, Wu, Winter, Hesse, Chen, Sigler, Litwin, Gray, Chess,
  Clark, Berner, McCandlish, Radford, Sutskever, and
  Amodei}]{brown2020language}
Tom Brown, Benjamin Mann, Nick Ryder, Melanie Subbiah, Jared~D Kaplan, Prafulla
  Dhariwal, Arvind Neelakantan, Pranav Shyam, Girish Sastry, Amanda Askell,
  Sandhini Agarwal, Ariel Herbert-Voss, Gretchen Krueger, Tom Henighan, Rewon
  Child, Aditya Ramesh, Daniel Ziegler, Jeffrey Wu, Clemens Winter, Chris
  Hesse, Mark Chen, Eric Sigler, Mateusz Litwin, Scott Gray, Benjamin Chess,
  Jack Clark, Christopher Berner, Sam McCandlish, Alec Radford, Ilya Sutskever,
  and Dario Amodei. 2020.
\newblock Language models are few-shot learners.
\newblock In \emph{Advances in Neural Information Processing Systems}, pages
  1877--1901.

\bibitem[{Caballero et~al.(2023)Caballero, Gupta, Rish, and
  Krueger}]{caballero2023broken}
Ethan Caballero, Kshitij Gupta, Irina Rish, and David Krueger. 2023.
\newblock Broken neural scaling laws.
\newblock In \emph{International Conference on Learning Representations}.

\bibitem[{Chen et~al.(2021)Chen, Tworek, Jun, Yuan, de~Oliveira~Pinto, Kaplan,
  Edwards, Burda, Joseph, Brockman, Ray, Puri, Krueger, Petrov, Khlaaf, Sastry,
  Mishkin, Chan, Gray, Ryder, Pavlov, Power, Kaiser, Bavarian, Winter, Tillet,
  Such, Cummings, Plappert, Chantzis, Barnes, Herbert-Voss, Guss, Nichol,
  Paino, Tezak, Tang, Babuschkin, Balaji, Jain, Saunders, Hesse, Carr, Leike,
  Achiam, Misra, Morikawa, Radford, Knight, Brundage, Murati, Mayer, Welinder,
  McGrew, Amodei, McCandlish, Sutskever, and Zaremba}]{chen2021evaluating}
Mark Chen, Jerry Tworek, Heewoo Jun, Qiming Yuan, Henrique~Ponde
  de~Oliveira~Pinto, Jared Kaplan, Harri Edwards, Yuri Burda, Nicholas Joseph,
  Greg Brockman, Alex Ray, Raul Puri, Gretchen Krueger, Michael Petrov, Heidy
  Khlaaf, Girish Sastry, Pamela Mishkin, Brooke Chan, Scott Gray, Nick Ryder,
  Mikhail Pavlov, Alethea Power, Lukasz Kaiser, Mohammad Bavarian, Clemens
  Winter, Philippe Tillet, Felipe~Petroski Such, Dave Cummings, Matthias
  Plappert, Fotios Chantzis, Elizabeth Barnes, Ariel Herbert-Voss,
  William~Hebgen Guss, Alex Nichol, Alex Paino, Nikolas Tezak, Jie Tang, Igor
  Babuschkin, Suchir Balaji, Shantanu Jain, William Saunders, Christopher
  Hesse, Andrew~N. Carr, Jan Leike, Josh Achiam, Vedant Misra, Evan Morikawa,
  Alec Radford, Matthew Knight, Miles Brundage, Mira Murati, Katie Mayer, Peter
  Welinder, Bob McGrew, Dario Amodei, Sam McCandlish, Ilya Sutskever, and
  Wojciech Zaremba. 2021.
\newblock \href {http://arxiv.org/abs/2107.03374} {Evaluating large language
  models trained on code}.
\newblock In \emph{arXiv}.

\bibitem[{Chollet(2019)}]{chollet2019measure}
François Chollet. 2019.
\newblock \href {http://arxiv.org/abs/1911.01547} {On the measure of
  intelligence}.
\newblock In \emph{arXiv}.

\bibitem[{Cobbe et~al.(2021)Cobbe, Kosaraju, Bavarian, Chen, Jun, Kaiser,
  Plappert, Tworek, Hilton, Nakano, Hesse, and Schulman}]{cobbe2021training}
Karl Cobbe, Vineet Kosaraju, Mohammad Bavarian, Mark Chen, Heewoo Jun, Lukasz
  Kaiser, Matthias Plappert, Jerry Tworek, Jacob Hilton, Reiichiro Nakano,
  Christopher Hesse, and John Schulman. 2021.
\newblock \href {http://arxiv.org/abs/2110.14168} {Training verifiers to solve
  math word problems}.
\newblock In \emph{arXiv}.

\bibitem[{Covert et~al.(2021)Covert, Lundberg, and Lee}]{covert2021explaining}
Ian~C. Covert, Scott Lundberg, and Su-In Lee. 2021.
\newblock Explaining by removing: a unified framework for model explanation.
\newblock \emph{The Journal of Machine Learning Research}.

\bibitem[{Finnveden(2020)}]{finnveden2020extra}
Lukas Finnveden. 2020.
\newblock \href
  {https://www.lesswrong.com/posts/k2SNji3jXaLGhBeYP/extrapolating-gpt-n-performance}
  {Extrapolating gpt-n performance}.

\bibitem[{Ganguli et~al.(2022{\natexlab{a}})Ganguli, Hernandez, Lovitt, Askell,
  Bai, Chen, Conerly, Dassarma, Drain, Elhage, El~Showk, Fort, Hatfield-Dodds,
  Henighan, Johnston, Jones, Joseph, Kernian, Kravec, Mann, Nanda, Ndousse,
  Olsson, Amodei, Brown, Kaplan, McCandlish, Olah, Amodei, and
  Clark}]{Ganguli2022}
Deep Ganguli, Danny Hernandez, Liane Lovitt, Amanda Askell, Yuntao Bai, Anna
  Chen, Tom Conerly, Nova Dassarma, Dawn Drain, Nelson Elhage, Sheer El~Showk,
  Stanislav Fort, Zac Hatfield-Dodds, Tom Henighan, Scott Johnston, Andy Jones,
  Nicholas Joseph, Jackson Kernian, Shauna Kravec, Ben Mann, Neel Nanda, Kamal
  Ndousse, Catherine Olsson, Daniela Amodei, Tom Brown, Jared Kaplan, Sam
  McCandlish, Christopher Olah, Dario Amodei, and Jack Clark.
  2022{\natexlab{a}}.
\newblock Predictability and surprise in large generative models.
\newblock In \emph{Conference on Fairness, Accountability, and Transparency}.
  ACM.

\bibitem[{Ganguli et~al.(2022{\natexlab{b}})Ganguli, Hernandez, Lovitt, Askell,
  Bai, Chen, Conerly, Dassarma, Drain, Elhage, El~Showk, Fort, Hatfield-Dodds,
  Henighan, Johnston, Jones, Joseph, Kernian, Kravec, Mann, Nanda, Ndousse,
  Olsson, Amodei, Brown, Kaplan, McCandlish, Olah, Amodei, and
  Clark}]{deep2022predict}
Deep Ganguli, Danny Hernandez, Liane Lovitt, Amanda Askell, Yuntao Bai, Anna
  Chen, Tom Conerly, Nova Dassarma, Dawn Drain, Nelson Elhage, Sheer El~Showk,
  Stanislav Fort, Zac Hatfield-Dodds, Tom Henighan, Scott Johnston, Andy Jones,
  Nicholas Joseph, Jackson Kernian, Shauna Kravec, Ben Mann, Neel Nanda, Kamal
  Ndousse, Catherine Olsson, Daniela Amodei, Tom Brown, Jared Kaplan, Sam
  McCandlish, Christopher Olah, Dario Amodei, and Jack Clark.
  2022{\natexlab{b}}.
\newblock Predictability and surprise in large generative models.
\newblock In \emph{ACM Conference on Fairness, Accountability, and
  Transparency}, pages 1747--–1764.

\bibitem[{Ghorbani et~al.(2021)Ghorbani, Firat, Freitag, Bapna, Krikun, Garcia,
  Chelba, and Cherry}]{ghorbani2021scaling}
Behrooz Ghorbani, Orhan Firat, Markus Freitag, Ankur Bapna, Maxim Krikun,
  Xavier Garcia, Ciprian Chelba, and Colin Cherry. 2021.
\newblock \href {http://arxiv.org/abs/2109.07740} {Scaling laws for neural
  machine translation}.
\newblock In \emph{arXiv}.

\bibitem[{Google(2024)}]{gemini15}
Google. 2024.
\newblock \href
  {https://blog.google/technology/ai/google-gemini-next-generation-model-february-2024/}
  {Gemini 1.5 blog}.

\bibitem[{Gordon et~al.(2021)Gordon, Duh, and Kaplan}]{gordon2021data}
Mitchell~A Gordon, Kevin Duh, and Jared Kaplan. 2021.
\newblock Data and parameter scaling laws for neural machine translation.
\newblock In \emph{Empirical Methods in Natural Language Processing}, pages
  5915--5922.

\bibitem[{He et~al.(2017)He, Liao, Zhang, Nie, Hu, and Chua}]{he2017neural}
Xiangnan He, Lizi Liao, Hanwang Zhang, Liqiang Nie, Xia Hu, and Tat-Seng Chua.
  2017.
\newblock Neural collaborative filtering.
\newblock In \emph{International Conference on World Wide Web}, pages 173--182.

\bibitem[{Hendrycks et~al.(2021)Hendrycks, Burns, Basart, Zou, Mazeika, Song,
  and Steinhardt}]{hendrycks2021measuring}
Dan Hendrycks, Collin Burns, Steven Basart, Andy Zou, Mantas Mazeika, Dawn
  Song, and Jacob Steinhardt. 2021.
\newblock Measuring massive multitask language understanding.
\newblock In \emph{International Conference on Learning Representations}.

\bibitem[{Henighan et~al.(2020)Henighan, Kaplan, Katz, Chen, Hesse, Jackson,
  Jun, Brown, Dhariwal, Gray, Hallacy, Mann, Radford, Ramesh, Ryder, Ziegler,
  Schulman, Amodei, and McCandlish}]{henighan2020scaling}
Tom Henighan, Jared Kaplan, Mor Katz, Mark Chen, Christopher Hesse, Jacob
  Jackson, Heewoo Jun, Tom~B. Brown, Prafulla Dhariwal, Scott Gray, Chris
  Hallacy, Benjamin Mann, Alec Radford, Aditya Ramesh, Nick Ryder, Daniel~M.
  Ziegler, John Schulman, Dario Amodei, and Sam McCandlish. 2020.
\newblock \href {http://arxiv.org/abs/2010.14701} {Scaling laws for
  autoregressive generative modeling}.
\newblock In \emph{arXiv}.

\bibitem[{Hernandez et~al.(2022)Hernandez, Brown, Conerly, DasSarma, Drain,
  El-Showk, Elhage, Hatfield-Dodds, Henighan, Hume, Johnston, Mann, Olah,
  Olsson, Amodei, Joseph, Kaplan, and McCandlish}]{hernandez2022scaling}
Danny Hernandez, Tom Brown, Tom Conerly, Nova DasSarma, Dawn Drain, Sheer
  El-Showk, Nelson Elhage, Zac Hatfield-Dodds, Tom Henighan, Tristan Hume,
  Scott Johnston, Ben Mann, Chris Olah, Catherine Olsson, Dario Amodei,
  Nicholas Joseph, Jared Kaplan, and Sam McCandlish. 2022.
\newblock \href {http://arxiv.org/abs/2205.10487} {Scaling laws and
  interpretability of learning from repeated data}.
\newblock In \emph{arXiv}.

\bibitem[{Hoffmann et~al.(2022)Hoffmann, Borgeaud, Mensch, Buchatskaya, Cai,
  Rutherford, de~las Casas, Hendricks, Welbl, Clark, Hennigan, Noland,
  Millican, van~den Driessche, Damoc, Guy, Osindero, Simonyan, Elsen, Vinyals,
  Rae, and Sifre}]{hoffmann2022an}
Jordan Hoffmann, Sebastian Borgeaud, Arthur Mensch, Elena Buchatskaya, Trevor
  Cai, Eliza Rutherford, Diego de~las Casas, Lisa~Anne Hendricks, Johannes
  Welbl, Aidan Clark, Tom Hennigan, Eric Noland, Katherine Millican, George
  van~den Driessche, Bogdan Damoc, Aurelia Guy, Simon Osindero, Karen Simonyan,
  Erich Elsen, Oriol Vinyals, Jack~William Rae, and Laurent Sifre. 2022.
\newblock An empirical analysis of compute-optimal large language model
  training.
\newblock In \emph{Advances in Neural Information Processing Systems}.

\bibitem[{Hu et~al.(2024)Hu, Liu, Han, Zhang, He, Zhao, Lin, Ding, Ou, Zeng,
  Liu, and Sun}]{hu2024predicting}
Shengding Hu, Xin Liu, Xu~Han, Xinrong Zhang, Chaoqun He, Weilin Zhao, Yankai
  Lin, Ning Ding, Zebin Ou, Guoyang Zeng, Zhiyuan Liu, and Maosong Sun. 2024.
\newblock Predicting emergent abilities with infinite resolution evaluation.
\newblock In \emph{International Conference on Learning Representations}.

\bibitem[{Ilić(2023)}]{ilić2023unveiling}
David Ilić. 2023.
\newblock \href {http://arxiv.org/abs/2310.11616} {Unveiling the general
  intelligence factor in language models: A psychometric approach}.

\bibitem[{Isik et~al.(2024)Isik, Ponomareva, Hazimeh, Paparas, Vassilvitskii,
  and Koyejo}]{isik2024scaling}
Berivan Isik, Natalia Ponomareva, Hussein Hazimeh, Dimitris Paparas, Sergei
  Vassilvitskii, and Sanmi Koyejo. 2024.
\newblock \href {http://arxiv.org/abs/2402.04177} {Scaling laws for downstream
  task performance of large language models}.
\newblock In \emph{arXiv}.

\bibitem[{Jethani et~al.(2022)Jethani, Sudarshan, Covert, Lee, and
  Ranganath}]{jethani2022fastshap}
Neil Jethani, Mukund Sudarshan, Ian Covert, Su-In Lee, and Rajesh Ranganath.
  2022.
\newblock \href {http://arxiv.org/abs/2107.07436} {Fastshap: Real-time shapley
  value estimation}.
\newblock In \emph{arXiv}.

\bibitem[{Kaplan et~al.(2020)Kaplan, McCandlish, Henighan, Brown, Chess, Child,
  Gray, Radford, Wu, and Amodei}]{kaplan2020scaling}
Jared Kaplan, Sam McCandlish, Tom Henighan, Tom~B. Brown, Benjamin Chess, Rewon
  Child, Scott Gray, Alec Radford, Jeffrey Wu, and Dario Amodei. 2020.
\newblock \href {http://arxiv.org/abs/2001.08361} {Scaling laws for neural
  language models}.
\newblock In \emph{arXiv}.

\bibitem[{Koren et~al.(2009)Koren, Bell, and Volinsky}]{koren2009matrix}
Yehuda Koren, Robert Bell, and Chris Volinsky. 2009.
\newblock Matrix factorization techniques for recommender systems.
\newblock \emph{Computer}, 42(8):30--37.

\bibitem[{Koren et~al.(2022)Koren, Rendle, and Bell}]{Koren2022}
Yehuda Koren, Steffen Rendle, and Robert Bell. 2022.
\newblock \emph{Advances in Collaborative Filtering}, pages 91--142. Springer.

\bibitem[{Liang et~al.(2023)Liang, Bommasani, Lee, Tsipras, Soylu, Yasunaga,
  Zhang, Narayanan, Wu, Kumar, Newman, Yuan, Yan, Zhang, Cosgrove, Manning,
  Ré, Acosta-Navas, Hudson, Zelikman, Durmus, Ladhak, Rong, Ren, Yao, Wang,
  Santhanam, Orr, Zheng, Yuksekgonul, Suzgun, Kim, Guha, Chatterji, Khattab,
  Henderson, Huang, Chi, Xie, Santurkar, Ganguli, Hashimoto, Icard, Zhang,
  Chaudhary, Wang, Li, Mai, Zhang, and Koreeda}]{liang2023holistic}
Percy Liang, Rishi Bommasani, Tony Lee, Dimitris Tsipras, Dilara Soylu,
  Michihiro Yasunaga, Yian Zhang, Deepak Narayanan, Yuhuai Wu, Ananya Kumar,
  Benjamin Newman, Binhang Yuan, Bobby Yan, Ce~Zhang, Christian Cosgrove,
  Christopher~D. Manning, Christopher Ré, Diana Acosta-Navas, Drew~A. Hudson,
  Eric Zelikman, Esin Durmus, Faisal Ladhak, Frieda Rong, Hongyu Ren, Huaxiu
  Yao, Jue Wang, Keshav Santhanam, Laurel Orr, Lucia Zheng, Mert Yuksekgonul,
  Mirac Suzgun, Nathan Kim, Neel Guha, Niladri Chatterji, Omar Khattab, Peter
  Henderson, Qian Huang, Ryan Chi, Sang~Michael Xie, Shibani Santurkar, Surya
  Ganguli, Tatsunori Hashimoto, Thomas Icard, Tianyi Zhang, Vishrav Chaudhary,
  William Wang, Xuechen Li, Yifan Mai, Yuhui Zhang, and Yuta Koreeda. 2023.
\newblock \href {http://arxiv.org/abs/2211.09110} {Holistic evaluation of
  language models}.
\newblock In \emph{arXiv}.

\bibitem[{Liu et~al.(2023)Liu, Lee, Jia, and Liang}]{liu2023question}
Nelson~F. Liu, Tony Lee, Robin Jia, and Percy Liang. 2023.
\newblock \href {http://arxiv.org/abs/2102.01065} {Do question answering
  modeling improvements hold across benchmarks?}
\newblock In \emph{arXiv}.

\bibitem[{Lundberg and Lee(2017)}]{lundberg2017unified}
Scott~M. Lundberg and Su-In Lee. 2017.
\newblock A unified approach to interpreting model predictions.
\newblock In \emph{International Conference on Neural Information Processing
  Systems}, pages 4768--4777.

\bibitem[{Muennighoff et~al.(2023)Muennighoff, Rush, Barak, Scao, Tazi, Piktus,
  Pyysalo, Wolf, and Raffel}]{muennighoff2023scaling}
Niklas Muennighoff, Alexander~M Rush, Boaz Barak, Teven~Le Scao, Nouamane Tazi,
  Aleksandra Piktus, Sampo Pyysalo, Thomas Wolf, and Colin Raffel. 2023.
\newblock Scaling data-constrained language models.
\newblock In \emph{Conference on Neural Information Processing Systems}.

\bibitem[{Nielsen(2016)}]{nielsen2016hierarchical}
Frank Nielsen. 2016.
\newblock \emph{Hierarchical Clustering}, pages 195--211. Springer.

\bibitem[{Ouyang et~al.(2022)Ouyang, Wu, Jiang, Almeida, Wainwright, Mishkin,
  Zhang, Agarwal, Slama, Ray, Schulman, Hilton, Kelton, Miller, Simens, Askell,
  Welinder, Christiano, Leike, and Lowe}]{ouyang2022training}
Long Ouyang, Jeffrey Wu, Xu~Jiang, Diogo Almeida, Carroll Wainwright, Pamela
  Mishkin, Chong Zhang, Sandhini Agarwal, Katarina Slama, Alex Ray, John
  Schulman, Jacob Hilton, Fraser Kelton, Luke Miller, Maddie Simens, Amanda
  Askell, Peter Welinder, Paul~F Christiano, Jan Leike, and Ryan Lowe. 2022.
\newblock Training language models to follow instructions with human feedback.
\newblock In \emph{Advances in Neural Information Processing Systems}, pages
  27730--27744.

\bibitem[{Owen(2024)}]{owen2024predictable}
David Owen. 2024.
\newblock \href {http://arxiv.org/abs/2401.04757} {How predictable is language
  model benchmark performance?}
\newblock In \emph{arXiv}.

\bibitem[{Perlitz et~al.(2024)Perlitz, Bandel, Gera, Arviv, Ein-Dor, Shnarch,
  Slonim, Shmueli-Scheuer, and Choshen}]{perlitz2024efficient}
Yotam Perlitz, Elron Bandel, Ariel Gera, Ofir Arviv, Liat Ein-Dor, Eyal
  Shnarch, Noam Slonim, Michal Shmueli-Scheuer, and Leshem Choshen. 2024.
\newblock \href {http://arxiv.org/abs/2308.11696} {Efficient benchmarking of
  language models}.
\newblock In \emph{arXiv}.

\bibitem[{Polo et~al.(2024)Polo, Weber, Choshen, Sun, Xu, and
  Yurochkin}]{polo2024tinybenchmarks}
Felipe~Maia Polo, Lucas Weber, Leshem Choshen, Yuekai Sun, Gongjun Xu, and
  Mikhail Yurochkin. 2024.
\newblock \href {http://arxiv.org/abs/2402.14992} {{tinyBenchmarks}: evaluating
  llms with fewer examples}.
\newblock In \emph{arXiv}.

\bibitem[{Ruan et~al.(2024)Ruan, Maddison, and
  Hashimoto}]{ruan2024observational}
Yangjun Ruan, Chris~J. Maddison, and Tatsunori Hashimoto. 2024.
\newblock \href {http://arxiv.org/abs/2405.10938} {Observational scaling laws
  and the predictability of language model performance}.
\newblock In \emph{arXiv}.

\bibitem[{Schaeffer et~al.(2023)Schaeffer, Miranda, and
  Koyejo}]{schaeffer2023are}
Rylan Schaeffer, Brando Miranda, and Sanmi Koyejo. 2023.
\newblock Are emergent abilities of large language models a mirage?
\newblock In \emph{Conference on Neural Information Processing Systems}.

\bibitem[{Shapley(1952)}]{shapley}
Lloyd~S. Shapley. 1952.
\newblock \href {https://doi.org/10.7249/P0295} {\emph{A Value for N-Person
  Games}}.
\newblock RAND Corporation.

\bibitem[{Shrivastava et~al.(2023)Shrivastava, Liang, and
  Kumar}]{shrivastava2023llamas}
Vaishnavi Shrivastava, Percy Liang, and Ananya Kumar. 2023.
\newblock \href {http://arxiv.org/abs/2311.08877} {Llamas know what gpts don't
  show: Surrogate models for confidence estimation}.
\newblock In \emph{arXiv}.

\bibitem[{Suzgun et~al.(2022)Suzgun, Scales, Schärli, Gehrmann, Tay, Chung,
  Chowdhery, Le, Chi, Zhou, and Wei}]{suzgun2022challenging}
Mirac Suzgun, Nathan Scales, Nathanael Schärli, Sebastian Gehrmann, Yi~Tay,
  Hyung~Won Chung, Aakanksha Chowdhery, Quoc~V. Le, Ed~H. Chi, Denny Zhou, and
  Jason Wei. 2022.
\newblock \href {http://arxiv.org/abs/2210.09261} {Challenging big-bench tasks
  and whether chain-of-thought can solve them}.
\newblock In \emph{arXiv}.

\bibitem[{Torregrossa et~al.(2020)Torregrossa, Claveau, Kooli, Gravier, and
  Allesiardo}]{torregrossa2020correlation}
Fran{\c{c}}ois Torregrossa, Vincent Claveau, Nihel Kooli, Guillaume Gravier,
  and Robin Allesiardo. 2020.
\newblock On the correlation of word embedding evaluation metrics.
\newblock In \emph{Language Resources and Evaluation Conference}, pages
  4789--4797.

\bibitem[{Wei et~al.(2022)Wei, Tay, Bommasani, Raffel, Zoph, Borgeaud,
  Yogatama, Bosma, Zhou, Metzler, Chi, Hashimoto, Vinyals, Liang, Dean, and
  Fedus}]{wei2022emergent}
Jason Wei, Yi~Tay, Rishi Bommasani, Colin Raffel, Barret Zoph, Sebastian
  Borgeaud, Dani Yogatama, Maarten Bosma, Denny Zhou, Donald Metzler, Ed~H.
  Chi, Tatsunori Hashimoto, Oriol Vinyals, Percy Liang, Jeff Dean, and William
  Fedus. 2022.
\newblock \href {http://arxiv.org/abs/2206.07682} {Emergent abilities of large
  language models}.
\newblock In \emph{arXiv}.

\bibitem[{Wei et~al.(2023)Wei, Wang, Schuurmans, Bosma, Ichter, Xia, Chi, Le,
  and Zhou}]{wei2023chainofthought}
Jason Wei, Xuezhi Wang, Dale Schuurmans, Maarten Bosma, Brian Ichter, Fei Xia,
  Ed~Chi, Quoc Le, and Denny Zhou. 2023.
\newblock \href {http://arxiv.org/abs/2201.11903} {Chain-of-thought prompting
  elicits reasoning in large language models}.
\newblock In \emph{arXiv}.

\bibitem[{Xiong et~al.(2023)Xiong, Liu, Molybog, Zhang, Bhargava, Hou, Martin,
  Rungta, Sankararaman, Oguz, Khabsa, Fang, Mehdad, Narang, Malik, Fan,
  Bhosale, Edunov, Lewis, Wang, and Ma}]{xiong2023effective}
Wenhan Xiong, Jingyu Liu, Igor Molybog, Hejia Zhang, Prajjwal Bhargava, Rui
  Hou, Louis Martin, Rashi Rungta, Karthik~Abinav Sankararaman, Barlas Oguz,
  Madian Khabsa, Han Fang, Yashar Mehdad, Sharan Narang, Kshitiz Malik, Angela
  Fan, Shruti Bhosale, Sergey Edunov, Mike Lewis, Sinong Wang, and Hao Ma.
  2023.
\newblock \href {http://arxiv.org/abs/2309.16039} {Effective long-context
  scaling of foundation models}.
\newblock In \emph{arXiv}.

\bibitem[{Yang et~al.(2019)Yang, Akimoto, Kim, and Udell}]{yang2019oboe}
Chengrun Yang, Yuji Akimoto, Dae~Won Kim, and Madeleine Udell. 2019.
\newblock Oboe: Collaborative filtering for automl model selection.
\newblock In \emph{Proceedings of the 25th ACM SIGKDD International Conference
  on Knowledge Discovery \& Data Mining}, page 1173–1183.

\bibitem[{Zhuocheng et~al.(2023)Zhuocheng, Gu, Zhang, and
  Feng}]{zhuocheng2023scaling}
Zhang Zhuocheng, Shuhao Gu, Min Zhang, and Yang Feng. 2023.
\newblock Scaling law for document neural machine translation.
\newblock In \emph{Findings of the Association for Computational Linguistics:
  EMNLP 2023}, pages 8290--8303.

\end{thebibliography}
\bibliographystyle{acl_natbib}

\appendix

\section{Pilot Demonstrations using Neural Collaborative Filtering}
\begin{figure*}[htb]
\centering
\includegraphics[width=\textwidth]{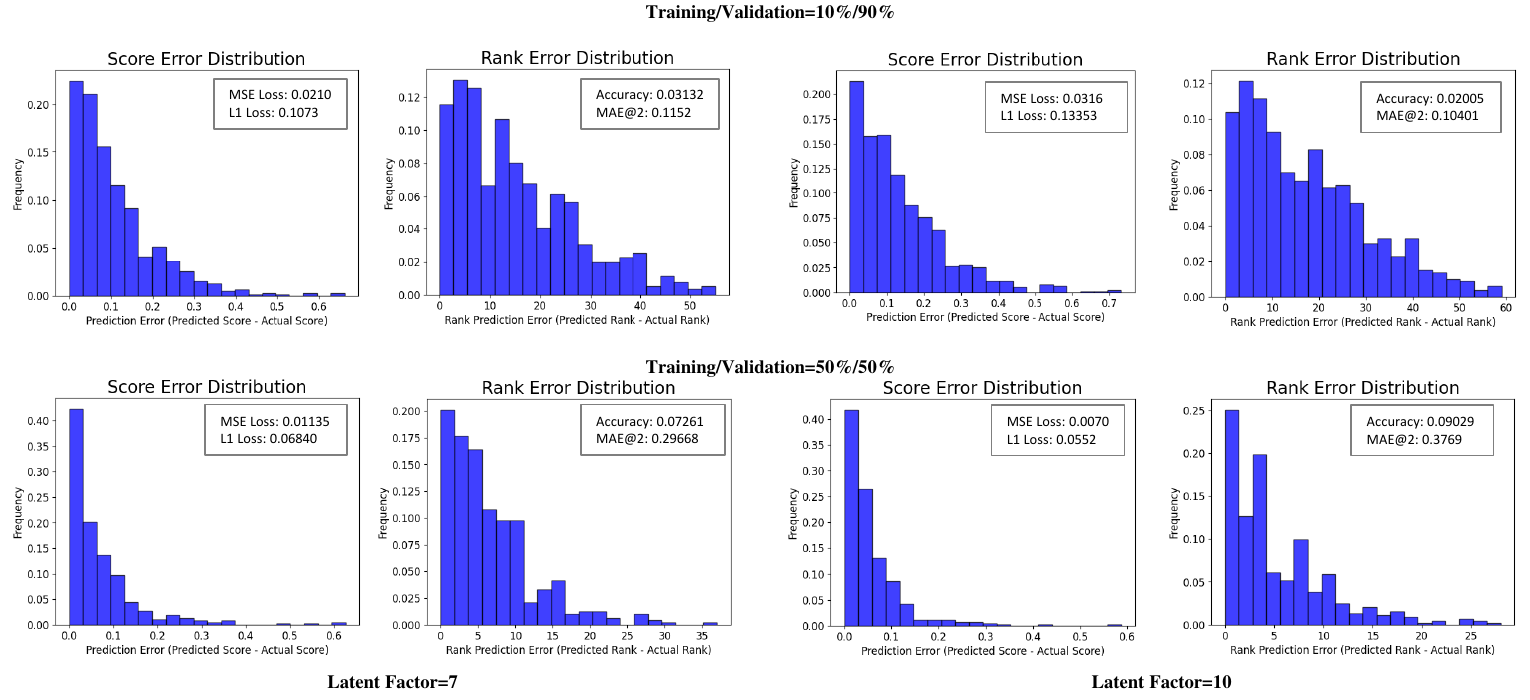}
\caption{\footnotesize \textbf{Error Distribution of Predictions (Normalized Score and Rank Derived by Score) Based on the HELM Lite Leaderboard Using Neural Collaborative Filtering:} We evaluate the effectiveness of Matrix Factorization (MF) using two latent factors, 7 and 10, across 2 training/validation split percentages. \textbf{Accuracy} is defined as the percentage of instances where the predicted rank equals the actual rank. \textbf{MAE@2} is defined as the percentage of instances where the absolute difference between the predicted rank and the actual rank is 2.}
\label{fig:pilot_ncf}
\end{figure*}

In this section, we supplemented the error distribution in Figure~\ref{fig:pilot_ncf}, which is generated using neural collaborative filtering on the HELM lite leaderboard. Compared to Figure~\ref{fig:pilot}, it is evident that neural collaborative filtering consistently outperforms MF across each setting.
\section{Collaborative Data}
\subsection{Data Description}

\paragraph{List of Models and Tasks.} 
The table~\ref{tab:descriptive_model_tasks} contains all the models and tasks we have collected.
\begin{table*}[htb]
\centering
\resizebox{0.98\textwidth}{!}{ 
\begin{tabular}{@{}cc@{}}
\hline
\textbf{Models} & \textbf{Tasks} \\
\midrule
\makecell{'LLama-2-7B', 'LLama-2-13B', 'LLama-2-70B', 'Llama 3 8B', 'Llama 3 70B',\\
       'GLM-130B', 'LLaMA-7B', 'LLaMA-13B', 'LLaMA-33B', 'LLaMA-65B',\\
       'GPT-3-175B', 'PaLM-540B', 'Claude-V3 Haiku', 'Claude-V3 Sonnet',\\
       'Claude-V3 Opus', 'GPT-4', 'gpt-3.5', 'BLOOM-176B', 'Luminous Base-13B',\\
       'Luminous Extended-30B', 'Luminous Supreme-70B', 'OPT-175B',\\
       'GPT-NeoX-20B', 'GPT-J-6B', 'sheared llama-2.7B', 'sheared llama-1.3B',\\
       'INCITE-Base-3B', 'INCITE-Base-7B', 'TinyLlama-1.1B', 'OpenLLaMA-3B-v1',\\
       'OpenLLaMA-3B-v2', 'Pythia-1.4B', 'Pythia-2.8B', 'Falcon-7B',\\
       'Falcon-40B', 'Falcon-180B', 'Mistral 7B', 'MPT-30B', 'MPT-7B',\\
       'chinchilla', 'Pythia-70M', 'Pythia-160M', 'Pythia-410M', 'Pythia-1B',\\
       'Pythia-6.9B', 'Pythia-12B', 'Gopher - 280B', 'Gopher - 44M',\\
       'Gopher - 117M', 'Gopher - 417M', 'Gropher - 1.4B', 'Gopher - 7.1B',\\
       'MT-NLG 530B', 'GLaM', 'Phi-1.5-1.3B', 'Phi-2-2.7B', 'Yi-6b', 'Yi-9b',\\
       'Baichuan 1-7B', 'Baichuan 1-13B-Base', 'Baichuan 2-7B-Base',\\
       'Baichuan 2-13B-Base', 'InternLM2-7B', 'InternLM2-20B', 'Skywork-13B',\\
       'BlueLM-7B', 'Qwen-7B', 'Qwen-14B', 'TigerBot-13b', 'TigerBot-70b',\\
       'Gemma-2b', 'Gemma-7b'}&\makecell{'BoolQ(0-shot)', 'BIG-bench hard(3-shot)','WinoGrande(0-shot)','WinoGrande(1-shot)',\\'Winogrande(5-shot)','PIQA(0-shot)','SIQA(0-shot)','HellaSwag(0-shot)','HellaSwag(10-shot)',\\'ARC-e','ARC-c(0-shot)','ARC-c(25-shot)','OBQA(zero-shot)','MMLU(5-shot)',\\'HumanEval(pass@1)','MBPP(3-shot)','GSM8K(4-shot)','MATH(4-shot)',\\'TriviaQA(5-shot)','NaturalQuestions(0-shot)','NaturalQuestions(1-shot)','NaturalQuestions(5-shot)',\\'NaturalQuestions(64-shot)','LAMBADA(0-shot)','AGIEval English (3-5 shot)','RACE-m',\\'RACE-h','LogiQA','WSC'}\\
\bottomrule
\end{tabular}
}
\caption{List of Models and Tasks}
\label{tab:descriptive_model_tasks}
\end{table*}

\paragraph{Description Factors for Models and Tasks}
\label{sec:description_features}
We have collected the characteristics of models and tasks in relevant aspects through model cards, technical reports, and academic papers. We have organized and introduced these characteristics, as well as the corresponding embedding methods, as listed in Table~\ref{tab:descriptive_feat}.
\begin{table*}[htb]
\centering
\resizebox{0.98\textwidth}{!}{ 
\begin{tabular}{@{}ccc@{}} 
\hline
\multicolumn{3}{c}{\textbf{Model}}  \\
\hline
\textbf{Factors} & \textbf{Description}   & \textbf{Embedding}   \\
\hline
Model Family  & Type of model family, \eg, \textsc{Llama 2}, \textsc{Pythia}  & Categorical Embedding \\
Pretraining Dataset Size (B)  & Data size in millions of tokens  & Numerical Embedding\\
Parameter Size (M)& Number of model parameters in millions &Numerical Embedding\\
GPUh  & GPU hours consumed   &     Numerical Embedding \\
FLOPs & Floating-point operations count  & Numerical Embedding \\
Context Window    & Max context size in tokens, \eg, 1024, 2048 & Categorical Embedding\\
Batch Size (M)    & Size of batches in millions,\eg, 1M, 2M      &    Categorical Embedding\\
Layers& Number of layers in the model    &   Numerical Embedding    \\
Number Heads      & Number of attention heads        &     Numerical Embedding\\
Key/Value Size    & Size of key/value in attention mechanism &  Numerical Embedding \\
Bottleneck Activation Size    & Size of activation in bottleneck layers &  Numerical Embedding \\
Carbon Emission (tCO2Eq)      & Carbon footprint of training     &  Numerical Embedding  \\ 
\hline
\multicolumn{3}{c}{\textbf{Task}} \\
\hline
Ability& Type of targeted cognitive ability, \eg, reasoning & Categorical Embedding \\
 TaskFamily   & Related task family ,\eg, ARC& Categorical Embedding\\
Output Format        & Format of task output, \eg, binary & Categorical Embedding\\
Few-Shot Setting   & Description of few-shot learning setting,\eg, zero-shot, 32-shot  &Categorical Embedding\\
\hline
\end{tabular}
}
\caption{Design Factors of Models and Tasks}
\label{tab:descriptive_feat}
\end{table*}

Note that during data collection, not all factors are available. For these missing factors, such as CO2 and GPU hours, we replace them as zero when entering data.

\subsection{Data Analysis}
\label{sec:data_analysis_appendix}
\begin{figure}[htb]
\begin{center}
\centerline{\includegraphics[width=0.5\textwidth]{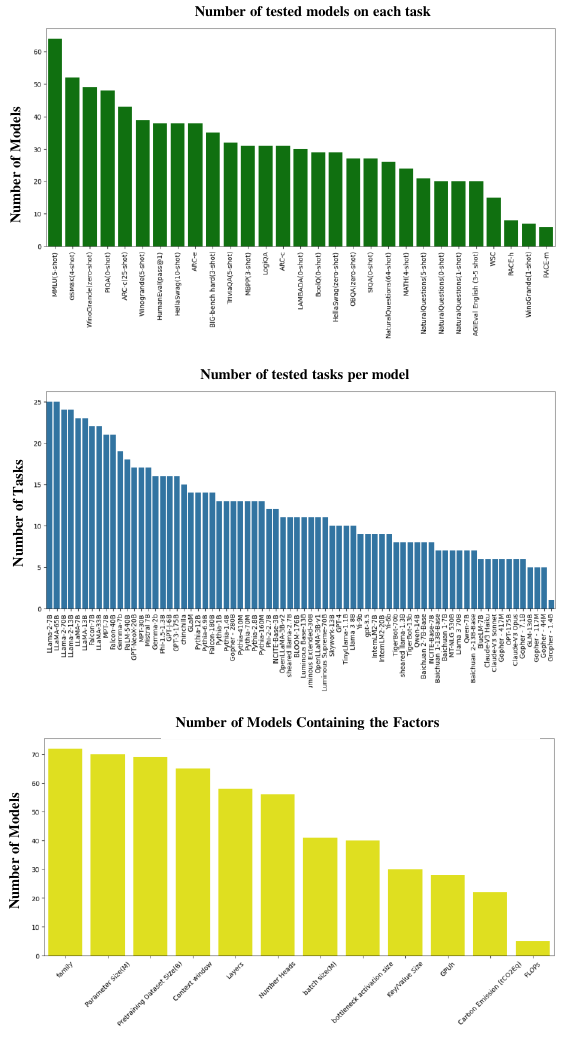}}
    \caption{The detailed distribution of collaborative data.}
\label{fig:detailed_analysis}
\end{center}
\vskip -0.2in
\end{figure}

We conducted a statistical analysis of the data we collected, specifically examining the number of models tested for each task, the number of tasks tested for each model, and the number of models described by each factor. Since each task is consistently associated with four factors, we did not create a distribution chart for this aspect.
\section{Experimental Setup}

\subsection{Evaluation Metrics}
\label{sec:appendix_evaluation}
Apart from visualization, we also evaluate the method based on two types of metrics: 1) \textsc{Score-Loss} Metric: we calculate \textsc{MSE Loss} and \textsc{L1 Loss} between predicted scores and true scores (normalized) on downstream tasks; 2) \textsc{Rank-Accuracy} Metric: researchers are sometimes not concerned with detailed scores but rather the rankings the model is in, so we calculate the accuracy of rank derived from the predicted scores, \textsc{Accuracy} and \textsc{MAE@2}. \textsc{Accuracy} refers to the percentage of instances where the predicted rank equals the true rank, and \textsc{MAE@2} refers to the percentage of instances where the absolute difference between the predicted rank and the true rank is in 2, the formulation as below:

\begin{equation}
    \text{Accuracy} = \left( \frac{\sum_{i=1}^{N} \mathbf{1}(r_i = \hat{r}_i)}{N} \right) \times 100\%,
\end{equation}

\begin{equation}
    \text{MAE@2} = \left( \frac{\sum_{i=1}^{N} \mathbf{1}(|r_i - \hat{r}_i| \leq 2)}{N} \right) \times 100\%,
\end{equation}
where $N$ is the total number of validation instances, $r_i$ is the true rank, $\hat{r}_i$ is the predicted rank derived by the predicted score; $\mathbf{1}(\cdot)$ is the indicator function that evaluates to 1 if the argument is true and 0 otherwise; $|\cdot|$ denotes the absolute value.

\subsection{Detailed Setting of Validation Prediction Accuracy Experiments }
\label{sec:validation_setting}
In this section, we detail the setup of each experiment in \ref{sec:validation}.
\paragraph{Different Prediction Methods.} Due to the 44\% sparsity of the collected collaboration matrix, we used 5\% of the known data as the validation set, with the remaining data serving as the observed training set. We trained each model five times through random splitting, deriving an average performance and variance. We configured our models with latent factors $=10$, learning rate $=0.01$, and iteration $=250,000$. The Figure~\ref{fig:different_models} is the results when random\_seed $=1$.
\paragraph{Predicting from Small to Large LMs.}
The focus here is on deriving the scaling law applicable to specific task metrics. Undeniably, traditional methods do not provide a directly usable scaling law across all downstream tasks for comparative analysis. However, we observed in the literature~\cite{ruan2024observational} that a sigmoidal curve with a single coefficient and a single bias value represents the scaling law for downstream tasks. Moreover, this curve's coefficients and bias values have a general range across all tasks, $w=[0.5,2],b=[-10,-3]$. Consequently, we set this range of coefficients and bias for this curve. Then we used the normalized scores of smaller models within the same model family and their corresponding parameter sizes to fit the scaling law curve for each task. This approach generally follows a ``pretrain-finetune'' methodology. Additionally, \textsc{CPP-2} refers to randomly selecting two scores from the observed performances of the model to be included in the training data. In this experiment, we use factor-enhanced NCF (setting is same as above).

\begin{table*}[!htbp]
\resizebox{1.0\textwidth}{!}{
\centering
\begin{tabular}{lccccc} 
\hline
\multirow{2}{*}{\textbf{Scaled LLMs}} & \multirow{2}{*}{\textbf{Prior Tasks}} & \multicolumn{2}{c}{\textbf{Score-Loss}} & \multicolumn{2}{c}{\textbf{Rank-Acc}} \\
\cline{3-6}
& & MSE Loss & Mean L1 Loss & Mean Prec.(\%) & MAE@2(\%) \\
\hline 
\multirow{2}{*}{\normalsize{LLaMA 2-70B}}  
& CF-0 & $1.34e^{-2}$ & $8.83e^{-2}$ & $16.7$ & $50.0$ \\
& CF-2 & $1.79e^{-2}\text{\small($1.3e^{-3}$)}$&$1.79e^{-2}$\text{\small($5.6e^{-4}$)} & $9.1$\text{\small($7.5e^{-3}$)} & $54.5$\text{\small($5.7e^{-4}$)}\\
\hline
 
\multirow{2}{*}{\normalsize{LLaMA 3-70B}}  
& CF-0 & $5.63e^{-2}$ & $19.27e^{-2}$ & $14.3$ & $71.4$ \\
& CF-2 & $1.7e^{-2}$\text{\small($1.41e^{-4}$)} & $10.7e^{-2}$ \text{\small($1.68e^{-3}$)} & $20.0$\text{\small($4.0e^{-2}$)}&$90.0$\text{\small($9.0e^{-2}$)} \\
\hline
 
\multirow{2}{*}{\normalsize{LLaMA-65B}}  
& CF-0 & $1.73e^{-2}$ & $9.78e^{-2}$ & $24.0$ & $80.0$\\
& CF-2 & $1.88e^{-2}$\text{\small($1.42e^{-5}$)} & $10.02e^{-2}$\text{\small($4.1e^{-4}$)} & $17.3$\text{\small($1.9e^{-3}$)}&$71.7$\text{\small($4.7e^{-4}$)}\\
\hline
 
\multirow{2}{*}{\normalsize{Luminous Supreme-70B}}  
& CF-0 & $6.06e^{-2}$ & $20.14e^{-2}$ & $27.27$ & $63.63$ \\
& CF-2 & $1.45e^{-2}$\text{\small($1.1e^{-5}$)} & $10.79e^{-2}$\text{\small($6.4e^{-7}$)} & $16.7$\text{\small($3.1e^{-3}$)}&$83.3$\text{\small($3.5e^{-3}$)} \\
\hline
 
\multirow{2}{*}{\normalsize{Pythia-12B}}  
& CF-0 & $2.19e^{-2}$ & $11.2e^{-2}$ & $21.42$ & $71.42$ \\
& CF-2 & $1.57e^{-2}$\text{\small($2.1e^{-6}$)} & $10.88e^{-2}$\text{\small($4.6e^{-8}$)} & $33.3$\text{\small($2.7e^{-2}$)} & $66.7$\text{\small($6.9e^{-3}$)} \\
\hline
 
\multirow{2}{*}{\normalsize{Yi-9b}}
& CF-0 & $3.20e^{-2}$ & $14.66e^{-2}$ & $44.4$ & $100.0$ \\
& CF-2 & $0.9e^{-2}$\text{\small($3.1e^{-4}$)} & $8.1e^{-2}$\text{\small($5.1e^{-6}$)} & $71.4\text{\small($9.1e^{-2}$)}$ & $100$\text{\small($0$)} \\
\hline

\multirow{2}{*}{\normalsize{Baichuan 2-13B-Base}}  
& CF-0 & $2.70e^{-2}$ & $12.84e^{-2}$ & $57.14$ & $100.0$ \\
& CF-2 & $1.0e^{-2}$\text{\small($4.9e^{-4}$)} & $7.5e^{-2}$\text{\small($4.7e^{-4}$)} & $40.0$\text{\small($6.2e^{-4}$)} & $100.0$\text{\small($0$)} \\
\hline

\multirow{2}{*}{\normalsize{Qwen-14B}}  
& CF-0 & $1.05e^{-2}$ & $7.96e^{-2}$ & $33.3$ & $100.0$ \\
& CF-2 & $3.1e^{-2}$\text{\small($1.8e^{-3}$)} & $11.1e^{-2}$\text{\small($6.6e^{-3}$)} & $25.0$\text{\small($7.1e^{-3}$)} & $91.7$\text{\small($6.9e^{-3}$)} \\
\hline
 
\multirow{2}{*}{\normalsize{TigerBot-70B}}  
& CF-0 & $8.02e^{-2}$ & $19.26e^{-2}$ & $12.5$ & $75.0$ \\
& CF-2 & $4.4e^{-2}$\text{\small($2.9e^{-6}$)} & $15.3e^{-2}$\text{\small($6.6e^{-5}$)} & $25.0$\text{\small($6.9e^{-3}$)} & $83.3$\text{\small($6.1e^{-3}$)} \\
\hline

\multirow{2}{*}{\normalsize{Gamma-7B}} 
& CF-0 & $4.94e^{-2}$ & $17.62e^{-2}$ & $15.79$ & $47.36$ \\
& CF-2 & $10.2e^{-2}$\text{\small($3.2e^{-5}$)} & $25.9e^{-2}$\text{\small($1.6e^{-4}$)} & $26.4$\text{\small($8.6e^{-4}$)} & $58.8$\text{\small($1.4e^{-2}$)} \\
\hline

\multirow{2}{*}{\normalsize{Falcon-180B}}  
& CF-0 & $5.00e^{-2}$ & $17.91e^{-2}$ & $14.58$ & $57.14$ \\
& CF-2 & $3.2e^{-2}$\text{\small($2.1e^{-5}$)} & $10.42e^{-2}$\text{\small($7.8e^{-5}$)} & $23.94$\text{\small($8.5e^{-2}$)} & $63.6$\text{\small($2.1e^{-5}$)}\\
\hline

\multirow{2}{*}{\normalsize{Gopher-280B}}  
& CF-0 & $14.48e^{-2}$ & $30.76e^{-2}$ & $15.38$ & $61.53$ \\
& CF-2 & $10.87e^{-2}$\text{\small($3.6e^{-5}$)} & $23.59$\text{\small($4.2e^{-4}$)} & $27.33$\text{\small($1.8e^{-3}$)} & $66.49$\text{\small($6.8e^{-3}$)}\\
\hline
\end{tabular}}
\caption{The accuracy of Predicting Scaled Large LMs in CPP-0, CPP-2. }
\label{tab:prediction_acc_scaled_model}
\end{table*}

\subsection{Detailed Setting of Analysis Experiments}
\label{sec:validation_setting_shapley}
\paragraph{Shapley-Value for Factor Importance Analysis.}
Given a predictive model \( f \) and a set of factors \( N \), the Shapley value of a factor \( i \) is computed as follows:

\begin{equation}
\begin{aligned}
\phi_i(v) 
&= \sum_{S \subseteq N \setminus \{i\}} \frac{|S|! (|N| - |S| - 1)!}{|N|!} \\
& \cdot \left[v(S \cup \{i\}) - v(S)\right], \\
\end{aligned}
\end{equation}
where:
\begin{itemize}[topsep=0pt,noitemsep,nolistsep,leftmargin=*]
\item \( N \) is the total set of factors.
\item \( S \) is a subset of factors excluding factor \( i \).
\item \( |S| \) is the number of factors in subset \( S \).
\item \( v(S) \) is the prediction model's output when only the factors in subset \( S \) are used.
\item \( v(S \cup \{i\}) \) is the model's output when the factors in subset \( S \) plus factor \( i \) are used.
\item The factorial terms \( |S|! \) and \( (|N| - |S| - 1)! \) weigh the contribution of each subset according to the number of factors included or excluded, ensuring a fair allocation across all possible combinations.
\end{itemize}

The Shapley value, \( \phi_i(v) \), quantifies the average marginal contribution of a factor \( i \) across all possible combinations of factors. The formula takes every subset \( S \) of the total factor set \( N \) that does not include \( i \), calculates the difference in the model's prediction output with and without factor \( i \) and averages this difference over all subsets. The averaging is weighted by the factor \(\frac{|S|! (|N| - |S| - 1)!}{|N|!}\), which corresponds to the number of permutations in which subset \( S \) appears as a prefix or suffix of the total set when factor \( i \) is added.

This approach ensures that each factor's contribution is assessed fairly and comprehensively, accounting for interactions with other factors and their unique impact when combined in different ways. Shapley values are particularly useful for factor importance analysis because they provide a solid theoretical foundation and are less biased than simpler importance metrics.

The Shapley value algorithm for analyzing feature (factor) importance is computationally intensive, which has led to the development of various approximation methods~\cite{jethani2022fastshap}. Fortunately, our predictive model involves a manageable number of factors, allowing us to use the most accurate direct computation method of Shapley values. Specifically, we apply an enumeration approach to compute Shapley values on a pre-trained factor-enhanced neural collaborative filtering model during the inference stage. This involves systematically masking factors to assess their impact.

For the implementation, we mask factors differently based on their data type as outlined in Table~\ref{tab:descriptive_feat}:
\begin{itemize}[topsep=0pt,noitemsep,nolistsep,leftmargin=*]
    \item \textbf{numerical factors}: we set the input factor values to zero;
    \item \textbf{categorical factors}: we set the corresponding embedding layer parameters to zero.
\end{itemize}

We then compute the difference in validation loss with and without each factor present, providing us with each factor's marginal contribution. This detailed approach allows us to quantify precisely how much each factor contributes to the model's predictive performance, providing valuable insights into factor importance and model behavior.
\section{Ablation Study}
\label{sec:ablation}

\subsection{Ablation on Sparsity Threshold}
\label{sec:ablation_sparsity}
\begin{figure}[!htbp]
\centering
\vspace{-10pt}
\includegraphics[width=0.45\textwidth]{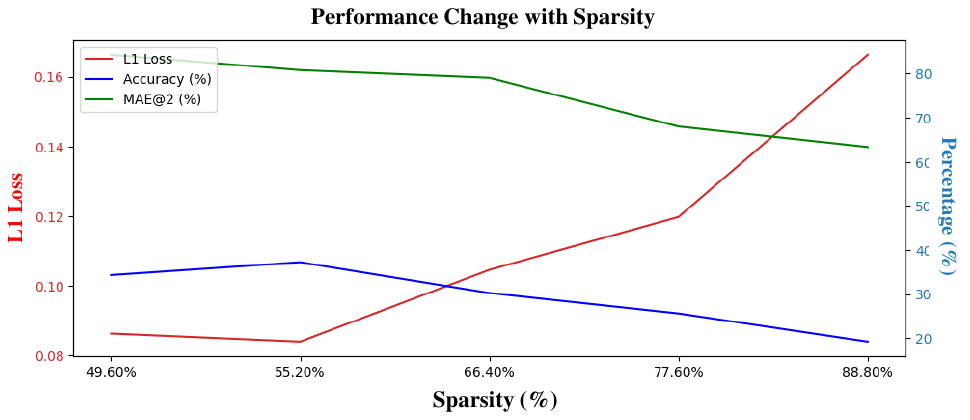}
\vspace{-10pt}
\caption{\footnotesize Relationship between matrix sparsity and three key performance metrics: L1 Loss, Accuracy, and MAE@2.}
\label{fig:sparsity}
\vspace{-10pt}
\end{figure}

To ascertain whether matrices composed of collaborative performance data can accurately predict the performance of LLMs, it is essential to consider the critical variable: the matrix \textbf{sparsity}. We assessed the impact of sparsity on prediction accuracy by manipulating the sparsity of the training matrix via masking. This method allowed us to obtain a reliable measure of average accuracy, as illustrated in Figure.~\ref{fig:sparsity}. It is noteworthy that our method of controlling sparsity only reduces the number of training samples. We ensured fairness in each comparative experiment by maintaining a consistent validation set throughout. 
During the experiment, we maintained the same settings for the learning rate and number of iterations as in the main experiment. 
To ensure the robustness of our experimental results, each reported outcome represents the average score after conducting five random splits.

The data we collected inherently has a sparsity of 44\%. Hence, we only have the remaining 46\% of collaborative data. As sparsity levels range from 49.60\% to 88.80\%(masking 10\% to 80\% of the collaborative data), the graph shows a pronounced increase in L1 Loss and a decrease in Accuracy, indicating deteriorating model performance with higher sparsity, especially when sparsity exceeds 60\%, where there is a significant drop in accuracy. Conversely, MAE@2 remains relatively stable before experiencing fluctuations, suggesting varying impacts on this metric.
Interestingly, accuracy even improves when sparsity reaches 50\%. We think the possible reason for this might be the presence of an optimal level of information reduction that removes redundant or noisy data without significantly compromising signal integrity. This phenomenon suggests that a moderate level of sparsity could potentially enhance model performance by focusing on more relevant factors.

\subsection{Ablation on Predicting Performance on Complex Reasoning and CoT Tasks}

From the model perspective, it is crucial for validating the feasibility of predictive methodologies to assess the predictive accuracy on special tasks potentially exhibiting ``emergent'' phenomena~\cite{suzgun2022challenging,wei2022emergent}, including complex reasoning and Chain of Thought (CoT) tasks~\cite{wei2023chainofthought}. ``Emergent' phenomena refers to the challenges associated with predicting performance from smaller models when the scale of a model expands significantly, resulting in discontinuous leaps in model capabilities. The existence of this phenomenon is subject to ongoing debate. Nonetheless, recent efforts~\cite{deep2022predict,hu2024predicting,owen2024predictable,ruan2024observational,schaeffer2023are} continue to focus on how scaling laws can be modified to mitigate the ``gap'' between smaller and larger models. This may involve modifying metrics or incorporating additional data points to linearize the growth curve or alternatively opting for a sigmoidal curve.

Theoretically, these challenges are not too difficult for our prediction method, as the underlying mechanism of ``emergent'' abilities reflects a type of similarity. This commonality manifests when models exceed a certain scale. By analyzing cross-model similarities—how other larger models demonstrate emergent capabilities compared to their smaller counterparts—we can enhance our predictive accuracy for the current model. Overall, these tasks are pivotal for comprehensive validation processes, \eg, \textsc{GSM8K}~\cite{cobbe2021training}, \textsc{BBH}~\cite{suzgun2022challenging}, \textsc{HumanEval}~\cite{chen2021evaluating} and \textsc{MBPP}~\cite{austin2021program}.

\begin{figure}[!htbp]
\centering
\includegraphics[width=0.45\textwidth]{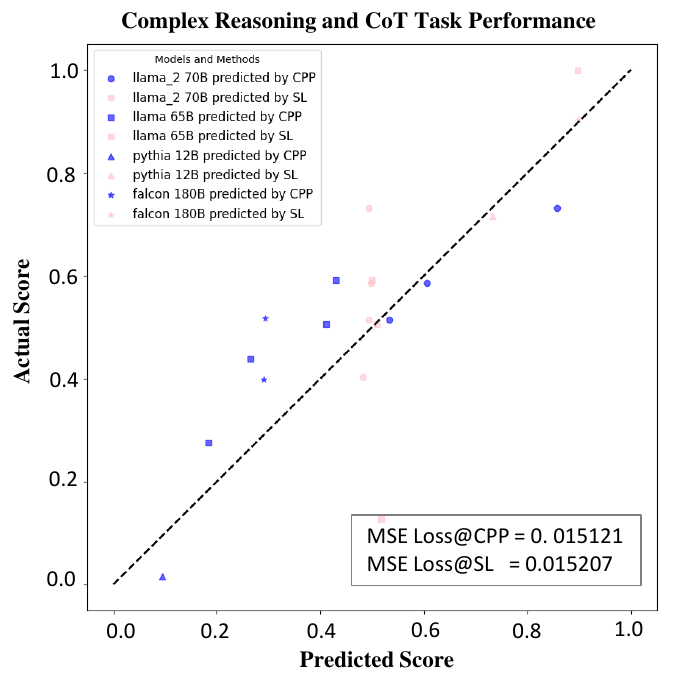}
\caption{\footnotesize Comparison of the predictive performance of collaborative performance prediction (CPP) versus traditional scaling laws (SL) for Large Language Models (LLMs) in Complex Reasoning and CoT Tasks.}
\label{fig:cot}
\end{figure}

In detail, if we want to evaluate the performance of predicting a model on these special tasks, the training data is the performance information from other model families, the smaller model of the same family, and the randomly selected two non-special tasks prior to the performance of this model. In our experiment, we tested the 4 models on these tasks, and then we plotted the test results on Figure~\ref{fig:cot}. As illustrated in Figure~\ref{fig:cot}, our predictive scores are more adaptive to each task, where the points are close along the ``perfect prediction'' line, which means our prediction method captures the similarity in the specific task across models. Our proposed method's MSE Loss is comparable to the scaling law, which shows the feasibility of \textsc{CPP} (in CPP-2).

\paragraph{Generalization to Completely New Tasks.} As presented in Tab.~\ref{tab:generalization}, \textsc{CPP-T0} and \textsc{CPP-T2} have a relative small error, demonstrating our method \textsc{CPP} shows reliable generalization. When \textsc{CPP-T2} has the prior performance of two models in this task, it has a significant drop compared to \textsc{CPP-T0}. These two experimental results inspire us that prediction and evaluation should be interactive, \ie, we should evaluate two small models or tasks to get true but low-cost results, and then the accuracy of prediction can be improved after obtaining the results.
\begin{table*}[!t]
\centering
\small 
\caption{The predictive performance (MSE) of CPP in the predictions of the completely new task. Here, \textsc{CPP-T0} refers to the predictive performance of CPP in the predictions of the completely new task, and \textsc{CPP-T2} refers to the predictive performance of CPP in the predictions of the task when we only know two models’ performance on this task, indicating CPP has no prior knowledge and few cases.}
\resizebox{0.9\textwidth}{!}{
\begin{tabular}{lcccc}
\toprule
\textbf{Models}  & \textbf{BoolQ(0-shot)} &	\textbf{BIG-bench hard(3-shot)}	&\textbf{HellaSwag(10-shot)}&	\textbf{HumanEval(pass@1)}\\
\midrule
\textbf{\textsc{CPP-T0}} &	0.02201 &	0.07103 &	0.03414  &	0.1244 \\
\textbf{\textsc{CPP-T2}} &	0.0182 &	0.00725	 &0.02506	 &0.0763 \\
\bottomrule
\end{tabular}
}
\label{tab:generalization}

\end{table*}

\subsection{Correlation between Models}
\label{sec:correlation_models}

\paragraph{Experiment.} We conducted a ``leave-one-out'' experiment to test the impact of Model A on the predictive performance of Model B. This involved masking Model A and using the performance of other models to train predictive methods, which were then validated on Model B to observe changes in loss. This approach generated a matrix with the masked model names on the X-axis and the validation model names on the Y-axis, with the values representing the change in loss.

The ``Leave-one-out'' experiment is a robust method commonly used in statistical analysis. To assess the impact of different models on the predictive performance of a specific model, we implemented a strategy where we systematically masked each selected model in the training set. The procedure involved masking each model individually and then training and testing the loss on a validation model. This process was repeated across all models, culminating in creating a matrix where axis=0 represents the masked model ID, and axis=1 represents the validation model ID. The values in the matrix correspond to the loss observed. This experiment was conducted under three different random seeds to ensure the stability and reliability of the results.

Subsequently, each model was used as a validation set, with the remaining data serving as the training set to calculate the loss for each model. This also resulted in a matrix where axis=1 indicates the validation model ID, and the columns[:, valid model id] represent the corresponding loss for that validation model. We derived a delta loss matrix by calculating the difference between these two matrices.

Given that each validation model has its own range of loss variations, we normalized the delta loss matrix. We then performed a row-based correlation analysis on this normalized matrix to assess each model's impact on predictive performance. The higher the correlation value between the two models, their effects on predictions are more similar.

\paragraph{Analysis.}  Based on this correlation matrix, we further conducted a hierarchical clustering analysis~\cite{nielsen2016hierarchical}. The results indicate that a set of models exists that are similar in their impact on the predictive performance of the model. Other models are far away from them. (Details in Table~\ref{tab:cluster})

This analysis not only helps us understand each model's specific contributions to predictive performance but also reveals the similarities and differences in functionality among the models, providing a crucial basis for model optimization and selection.

\begin{figure*}[!htbp]
\centering
\includegraphics[width=0.98\textwidth]{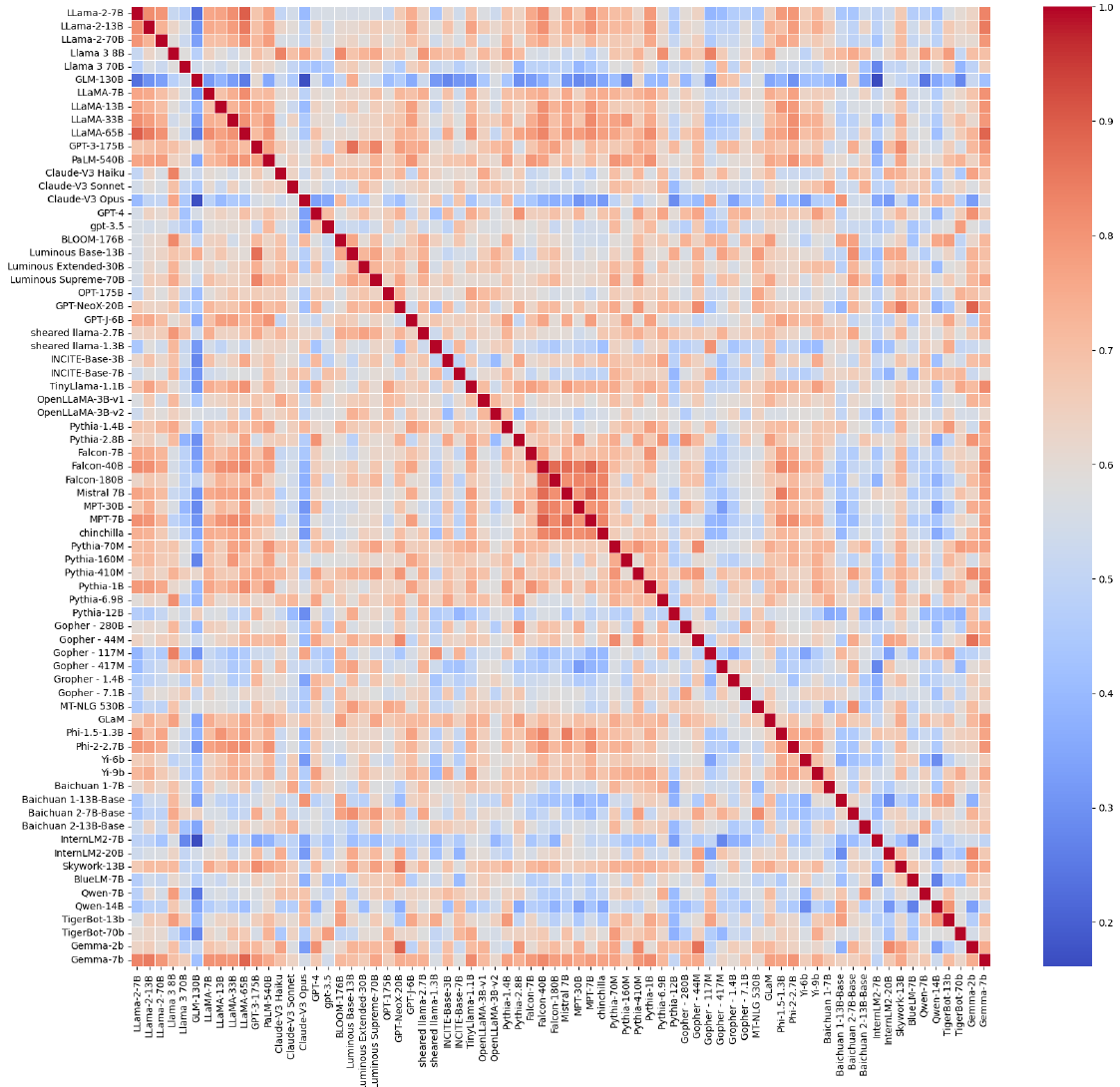}
\caption{\footnotesize The correlation heatmap of impacts of different models on prediction performance. 
}
\label{fig:correlation_models}
\vskip -0.3in
\end{figure*}

We performed a row-wise correlation analysis~\ref{fig:correlation_models} on this matrix and discovered that models from the same family tend to have similar impacts on predictions, as do models of the same size. After conducting a hierarchical distance analysis, we concluded that a group of models exists that, when more performance data is available, can significantly enhance the accuracy of the predictive models. There are also what might be termed ``noise model performances'' in our analysis~\ref{sec:correlation_models}.

\begin{figure}[!htbp]
\begin{center}
\centerline{\includegraphics[width=0.45\textwidth]{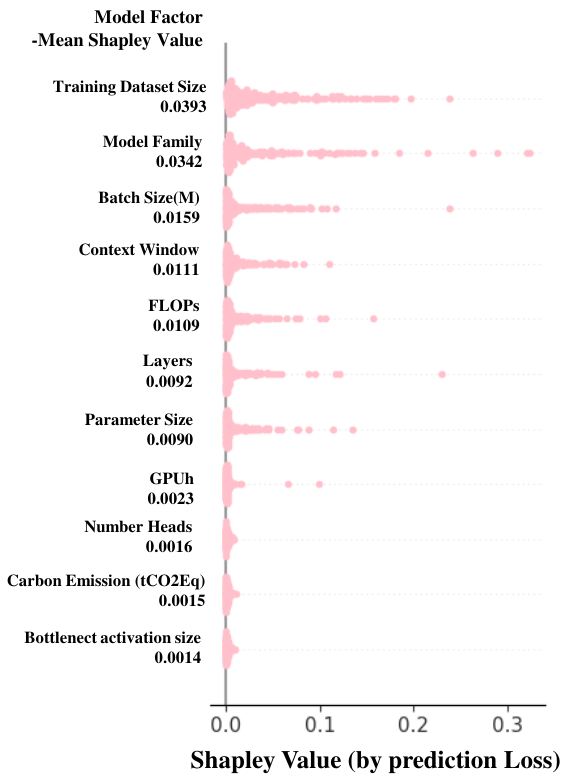}}
\caption{\footnotesize Instance Distribution of the model factor Shapley value. $X$-axis represents the Shapley value, which indicates the degree of prediction loss change; $Y$-axis indicates the factor names in order of importance from top to bottom. Each point represents an instance.}
\label{fig:model_shapley}
\end{center}

\end{figure}

\begin{figure}[!htbp]
\centering
\includegraphics[width=0.45\textwidth]{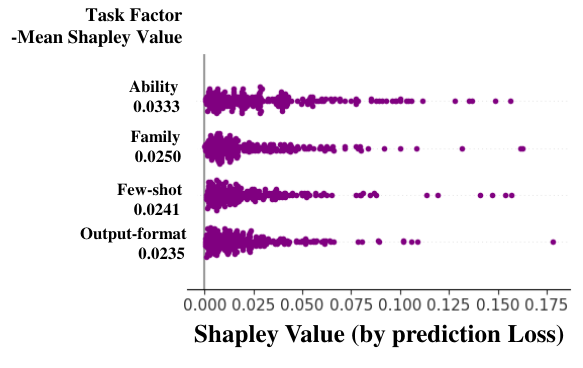}
\caption{\footnotesize Instance Distribution of the task factor Shapley value. $X$-axis represents the Shapley value, which indicates the degree of prediction loss change; $Y$-axis indicates the factor names in order of importance from top to bottom. Each point represents an instance.}
\label{fig:task_shapley}

\end{figure}

\subsection{Correlation between Tasks}

\begin{figure*}[!htbp]
\centering
\includegraphics[width=0.98\textwidth]{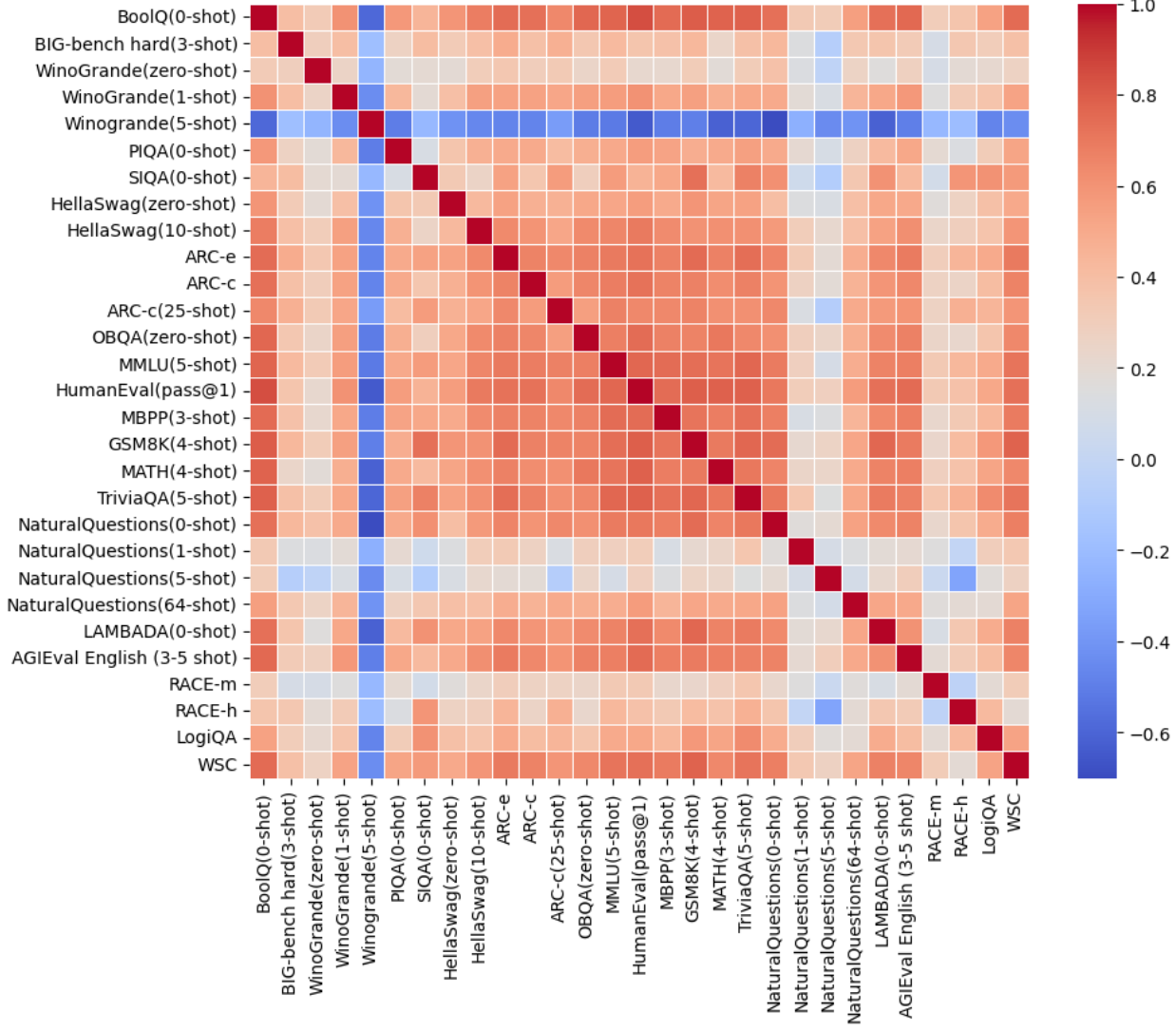}
\caption{\footnotesize The correlation heatmap of impacts of different tasks on prediction performance.}
\label{fig:correlation_tasks}
\vskip -0.3in
\end{figure*}

We also conducted ``leave-one-out'' experiments on these tasks and created a heatmap figure.~\ref{fig:correlation_tasks} of the correlations. Tasks with similar targeted ability testing capabilities demonstrated similar influences, such as \textsc{GSM8K}, \textsc{MATH}~\cite{hendrycks2021measuring}, \textsc{ARC}~\cite{chollet2019measure}, and \textsc{HumanEval}, which all require complex reasoning abilities.

\section{Others}
\subsection{Visualization}
\begin{figure*}[!htbp]
\centering
\includegraphics[width=0.98\textwidth]{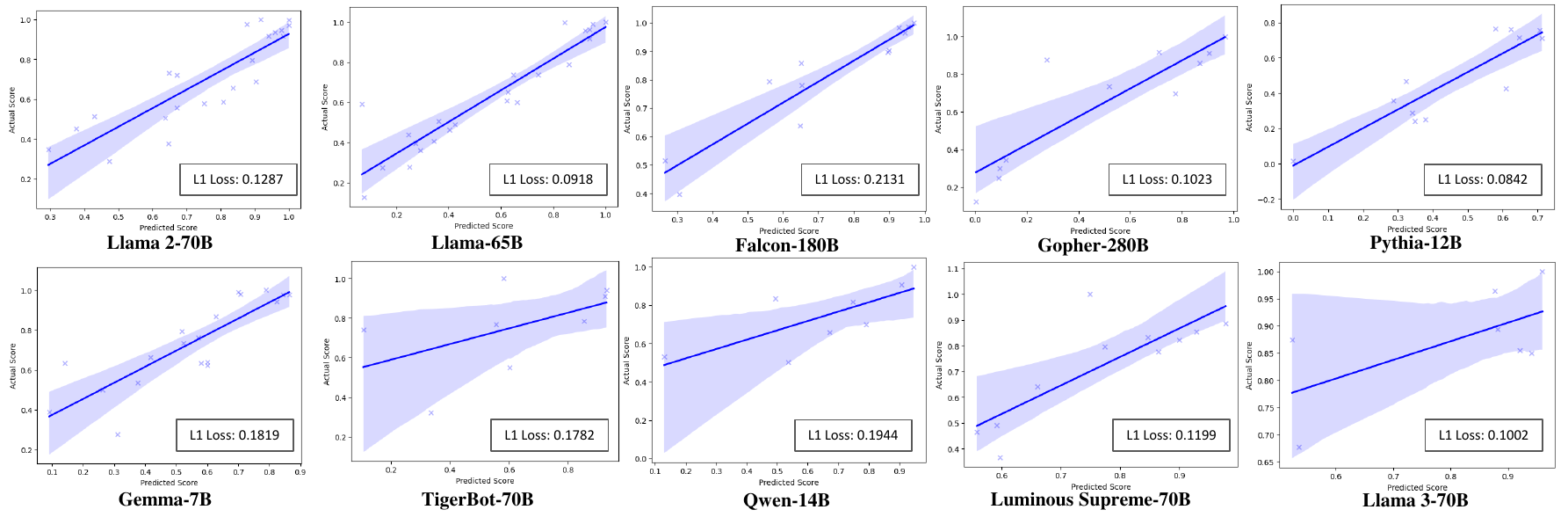}
\caption{\footnotesize Prediction performance of various scaled Language Models on downstream tasks. This figure illustrates regression plots comparing the predicted versus actual performance normalized scores for a series of large language models, including Llama-2-70B, Llama-65B, Falcon-180B, Gopher-280B, Pythia-12B, Gemma-7B, TigerBot-70B, Qwen-14B, Luminous Supreme-70B, and Llama-3-70B. Each subplot displays a regression line with a shaded 95\% confidence interval and includes the L1 loss for each model's predictions, highlighting the accuracy and variability of predictive capabilities across different models. 
}
\label{fig:visualization}
\vskip -0.3in
\end{figure*}

The figure~\ref{fig:visualization} is the visualization for the prediction performance of scaled language models on downstream tasks.

\begin{table*}[htb]
\centering
\resizebox{0.98\textwidth}{!}{
\begin{tabular}{c|c}
\hline
\textbf{Distance Cluster} & \textbf{Models} \\
\hline
1 & \makecell{LLama-2-7B, LLama-2-13B, LLama-2-70B, Llama 3 8B, \\LLaMA-7B, LLaMA-65B, Claude-V3 Haiku, Claude-V3 Sonnet,\\ Claude-V3 Opus, GPT-4, BLOOM-176B, Luminous Extended-30B,\\ Luminous Supreme-70B, OPT-175B, GPT-NeoX-20B, sheared llama-2.7B,\\ sheared llama-1.3B, INCITE-Base-3B, INCITE-Base-7B, OpenLLaMA-3B-v1, Pythia-1.4B,\\ Pythia-2.8B, Pythia-70M, Pythia-410M, Pythia-6.9B,\\ Gopher - 280B, Gopher - 44M, Gopher - 117M, MT-NLG 530B, GLaM,\\ Baichuan 1-7B, Baichuan 1-13B-Base, Baichuan 2-7B-Base, Baichuan 2-13B-Base,\\ Skywork-13B, Qwen-7B, Qwen-14B, TigerBot-13b,\\ Gemma-2b, Gemma-7b} \\
\hline
2 & gpt-3.5, Falcon-7B, Pythia-1B, Gropher - 1.4B, Yi-9b, TigerBot-70b \\
\hline
3 & LLaMA-33B \\
\hline
4 & Yi-6b \\
\hline
5 & BlueLM-7B \\
\hline
6 & Falcon-40B \\
\hline
7 & MPT-7B \\
\hline
8 & Falcon-180B \\
\hline
9 & PaLM-540B \\
\hline
10 & Pythia-160M \\
\hline
11 & GPT-J-6B \\
\hline
12 & GPT-3-175B, Luminous Base-13B \\
\hline
13 & Gopher - 417M \\
\hline
14 & Llama 3 70B \\
\hline
15 & LLaMA-13B \\
\hline
16 & TinyLlama-1.1B \\
\hline
17 & Phi-1.5-1.3B \\
\hline
18 & Gopher - 7.1B \\
\hline
19 & InternLM2-20B \\
\hline
20 & GLM-130B \\
\hline
21 & MPT-30B \\
\hline
22 & chinchilla \\
\hline
23 & Mistral 7B \\
\hline
24 & InternLM2-7B \\
\hline
25 & OpenLLaMA-3B-v2 \\
\hline
26 & Phi-2-2.7B \\
\hline
27 & Pythia-12B \\
\hline
\end{tabular}
}
\caption{Distance Cluster of Models}
\label{tab:cluster}
\end{table*}

\end{document}